# Using Large Language Models to Create AI Personas for Replication, Generalization and Prediction of Media Effects: An Empirical Test of 133 Published Experimental Research Findings

Leo Yeykelis[1*] Kaavya Pichai[2] James J. Cummings[3] Byron Reeves[2]

[1]Viewpoints AI [2]Stanford University [3]Boston University
*Corresponding author: leo@viewpoints.ai

**ABSTRACT**

This report analyzes the potential for large language models (LLMs) to expedite accurate replication and generalization of published research about message effects in marketing. LLM-powered participants (personas) were tested by replicating 133 experimental findings from 14 papers containing 45 recent studies published in the Journal of Marketing. For each study, the measures, stimuli, and sampling specifications were used to generate prompts for LLMs to act as unique personas. The AI personas, 19,447 in total across all of the studies, generated complete datasets and statistical analyses were then compared with the original human study results. The LLM replications successfully reproduced 76% of the original main effects, demonstrating strong potential for AI-assisted replication. The overall replication rate including interaction effects was 68%. Furthermore, a test of how human results generalized to different participant samples, media stimuli, and measures showed that replication results can change when tests go beyond the parameters of the original human studies. Implications are discussed for the replication and generalizability crises in social science, the acceleration of theory building in media and marketing psychology, and the practical advantages of rapid message testing for consumer products.

## Study Overview and Related Work

Research about the effectiveness of media messages is increasingly difficult, attributable to administrative challenges (e.g., stimulus acquisition and creation, data management demands for digital trace data, and acquisition of participants, especially for special audiences like children, minorities and international groups), as well as requirements to deal with new and critical challenges to the very nature of social research. The latter challenges include existential issues of replication and reproducibility, and the ability to generalize findings beyond the details of original studies, including to different people, media stimuli, and experimental contexts. We briefly review these issues with an eye toward our current test of whether new LLM tools may help solve the problems mentioned, and with an interest in significant research advantages in cost, time, and research personnel. The fundamental question for this project is whether an AI can accurately answer research questions, with accuracy measured in this case as the ability of AI personas to replicate the direction and magnitude of effects in AI studies that are identical to the human studies in all other qualities (i.e., the same media stimuli, measures, and participant sample specifications).

### *AI Proposals for Social Research*

The application of AI models to social research has been discussed extensively since LLMs became readily available, and generally with optimism that LLMs might be a powerful new tool with which to study human behavior in a variety of disciplines. There is growing interest in the prospect of leveraging AI to address historical and ongoing crises in social research, and initial attempts to use AI in research replication have shown promise (Bail, 2024). Recent advancements in LLMs have expanded applications to common research tasks, initially including literature review automation, hypothesis generation, and scientific discovery assistance (Wang et al., 2023).

More recent studies have had the explicit goal of comparing AI and human study results to test the accuracy of LLMs against the ground truth of prior identical human studies. These studies explore the prospect that LLMs could be used to replicate human research about social behavior, augmenting confidence in human study results, and possibly even replacing them. Argyle, et al., 2023, compared AI and human samples across a variety of national surveys, showing that “silicon” samples were capable of nuanced and complex representations of attitudes and ideas. These authors were able to show that proper conditioning of LLMs could mitigate known AI algorithmic biases (e.g., with respect to race and gender) which have been a concern about the use of LLMs in social research since the use of AI models were first proposed. Most recently, Park, et al., (2024) showed that general-purpose computational agents could replicate human responses on the General Social Survey 85% as accurately as human participants replicate their own answers. Several other reports discuss the needed research infrastructures to use AI in social and computational research, with particular attention to best practices in the construction of prompts, and methods to maximize the success (defined as agreement with human studies) of AI applications that involve classification, free-form coding and annotations (e.g., Ziems, et al., 2024). Not all evaluations are positive, however, about the prospects of AI to replace humans in social research (e.g., Harding, et al., 2023; Rosi, et al., 2024), but it is fair to conclude that transformational prospects are sufficiently interesting that a large number of research groups, often collaborating across technical and social science disciplines, are enthusiastic enough to keep exploring.

The enthusiasm about AI generally has led to focused applications in psychology. These efforts are significant for marketing because the stimuli, measures and analytics are similar to research in media psychology and consumer behavior. Binz et al. (2023) showed, for example, that GPT-3 could replicate psychological experiments, finding success particularly in research about language-based tasks. Cui, et al., (2024) showed that 76% of 618 main effects from psychology studies could be replicated, albeit with diminished effect sizes relative to the original research. And perhaps the most thorough description of an infrastructure for using LLMs in psychology was presented in the study by Argyle, et al., (2023). They introduced a critical metric for evaluating the success of LLMs in social research, called algorithmic fidelity, which allowed evaluation of whether an LLM could be prompted, using thousands of sociodemographic backstories from real human participants, to represent the same ideas and attitudes that were found in large national surveys. Their conclusion was that an LLM could indeed pass a social science version of the Turing Test; namely, that AI could generate data patterns with a high correspondence to human responses in areas relevant to psychology.

***Applications and Tests of AI in Marketing Science***

Applications of LLMs in marketing science have also now begun, initially including demonstrations of LLM uses, for example, to assess sentiment in consumer reviews (Zhang et al., 2022) and to generate marketing content (Kshetri et al., 2024). The potential of AI to replicate or even replace experimental studies, however, has remained largely unexplored until recently. Two new studies provided notable and useful guidance for this project.

One study (Arora, et al., 2024) showed that LLMs can be used as collaborators in hybrid marketing research that uses both AI and human input. They demonstrated that LLMs can generate synthetic respondents that provide and moderate in-depth qualitative interviews, and in cases when an AI collaborates with a human to conduct and interpret interviews, the results are rated by human judges as more useful and skilled than when either AI or humans conduct the interviews alone. The same study also reported success with AI-human collaboration on quantitative assessments, showing that few-shot training of LLMs (most importantly including the results of qualitative interviews related to the same research questions) could replicate the direction of responses to quantitative measures previously collected in a human study of attitudinal and purchasing responses to a retail product. Few-shot training with the same data improved the results even more, and particularly with respect to increasing the heterogeneity of responses across respondents.

In another marketing study, Li, et al., (2024) conducted a detailed test of the ability of LLMs to generate text responses to questions about retail products (products included cars, apparel and hotel brands). The human data that were compared to new AI tests were from surveys about the brands, and the data were used to create perceptual maps of brand attributes and evaluations. The study reported agreement between LLM-generated perceptual maps and human subject maps for over 75% of the comments generated. Perceptual maps for car brands showed better agreement with the LLM-generated data than for apparel and hotel brands, suggesting that the use of LLMs might work better for high-involvement products like automobiles. These results provide good support for using LLMs to substitute for at least some form of market research (perceptual surveys and maps), and they provide encouragement to extend the comparison of LLM and human results to other research designs, including generating quantitative responses to specific brand and product messages, the aim of this project.

***Extending the Uses of LLMs in Marketing Science***

The purpose of the present study was to test the usefulness of LLM models with respect to three different success criteria in research: (1) replication of empirical studies that test ideas about how humans respond to specific marketing messages; (2) generalization of empirical findings about message effects beyond the parameters of original studies; and (3) an evaluation of research efficiency defined as the cost, speed and administrative ease of incorporating research into decisions, both those in professional settings and with respect to the speed of theory development in marketing science.

*Replication*. The replication crisis across the sciences has received considerable attention. Ioannidis (2005) suggested that most published research findings are false, citing small sample sizes, small effect sizes, and researcher degrees of freedom as contributing factors. The Open Science Collaboration (2015) corroborated this concern, successfully replicating only 39% of psychology studies. Munafò et al. (2017) identified publication bias, p-hacking, and lack of replication incentives as key factors in the crisis. Summaries of research biases that cause poor replication have taken a prominent places in organized attempts, via conferences and edited volumes, to reform what many believe is the most serious crisis related to research accuracy and integrity in the social sciences (e.g., Jussim, Krosnick & Stevens, 2022).

The importance of replication has often been noted in the marketing literature (e.g., Hubbard & Armstrong, 1994; Easley, et al., 2000) and attempts at replication now appear in several business literatures. For example, Motoki and Iseki (2022) found that only two of ten influential studies about sensory marketing could be replicated. Davis, et al., (2023) were able to replicate only six of ten prominent experiments in operations management that had been published in *Management Science*. Two of the replications showed mixed results and two studies could not be replicated.

As we previously noted, most of the attempts to compare AI and human studies in marketing have focused on one or a just a few specific journal studies or on proprietary data, and not on replication of large collections of published results, as is the goal of the present study. We believe the prospects for AI replication are critical to examine. If replication is indeed a cornerstone of science, then methods will need to be developed, possibly including AI personas, that take considerably less time and resources, and that offer a process for researchers to complete replications that will not divert substantial effort from other research activities.

*Generalization***.** The crisis of replication has been extended importantly to the generalizability of results. Yarkoni (2022) comprehensively reviewed the current state of generalization issues in social science and argued that researchers quite often draw overly broad conclusions from narrow empirical findings. This critique is particularly pertinent in marketing and media effects research, where findings are frequently presumed to generalize across diverse consumer contexts. The importance of generalization issues, similar to replication, are increasingly prominent in the business literature (e.g., Hulland & Huston, 2018; introduction to special issue of *Journal of the Academy of Marketing Science* on generalizations in marketing). Many of the suggestions about how to test generalizability, however, involve literature reviews and meta-analyses. Yarkoni's solutions (2022) also included more rigorous consideration of research boundary conditions, increased use of large-scale naturalistic datasets, and development of computational tools for comprehensive analysis of research claims. But like replication, it is quite difficult to conduct new empirical tests that will show if and how results might extend to different people, messages, research contexts and measures. In the AI literature so far, almost all uses of AI concern validation of human studies using the exact parameters of the original studies, but it may be equally interesting, and perhaps even more important for theory and practice, to consider how AI studies could be used to efficiently *elaborate* findings beyond initial published specifications. If successful, those tests could provide critical evidence about the boundaries for generalizing results to similar but not identical conditions.

*Research Efficiency*. An obvious promise of AI solutions for replicability and generalizability problems is that the necessary extra research burden (i.e., the need to gather more and different data before acceptance of results) can be lessened with efforts that take only a small fraction of the time and money to complete compared to what would be needed if the extensions all involved human subjects and human researchers. Efficiencies of time and money would indeed be welcome, but to those advantages we add an *intellectual* value of research speed that would benefit everything from theory building to message and intervention designs. In professional practice, business strategy, and the design of products and services, AI might allow results to be used at a pace that would allow design and testing to be more iterative. If the use of a research result for theory or design depends on confirmation of its accuracy, or if the usefulness of a result depends on whether it extends to different samples of people or stimuli, those elaborations might be tested within minutes and hours, rather than put on hold until human confirmatory studies could be completed (if ever). In theory building, valuable science depends on an inductive-deductive cycle that uses descriptive data to generate and test theories. That cycle often occurs in months and years, but if the process could be speeded, descriptions might be more tightly coupled with the formulation of hypotheses, dramatically accelerating the development of new theory.

***The present study***

The studies reviewed above have set the table for larger projects that test at scale applications of LLMs to marketing science. While interest grows in addressing the replication crisis and leveraging AI in research, empirical efforts have not extensively explored AI's capacity to replicate complex research scenarios where users respond to specific textual *and* visual stimuli. Our study aims to address this gap by systematically applying LLM-powered participants to replicate recent marketing experiments that investigate human responses to media stimuli, assessing the feasibility of this approach, and potentially enhancing the reliability and generalizability of research findings in marketing.

In summary, this study makes several important new contributions to the exploration of AI options in marketing research. We believe our study is the first project to accomplish the following: (1) the application of AI personas across multiple (n=45) published studies (all current AI studies focus on only one or two datasets); (2) the first to concentrate exclusively on quantitative data (rather than emphasis on qualitative responses) that includes statistical estimates of the magnitude of experimental differences rather than just directional predictions; (3) the first to focus on relationships between variables rather than univariate estimates (e.g., estimates of survey response distributions or estimates of perceptual maps); (4) the first to use multi-modal models (words *and* pictures) to test replications and generalizations; (5) the first to use actual visual stimuli from the studies being replicated (all other marketing studies use text stimuli or text that describes visuals); (6) the first to exactly match the sampling parameters of AI replications to the human samples reported in the original publications; (7) the first to use the complete set of exact measures employed in the original human studies (rather than example questions from indexes); (8) the first to concentrate on zero-shot (rather than few-shot) models and prompts that could more quickly accelerate theory building and message design; and (9) the first to extend the use of AI to test the *generalization* of results to new groups of people, new media examples, similar measures, and different research contexts (e.g., online surveys vs. laboratory experiments, different participant recruiting methods).

***Benchmarking and Analytic Strategies***

A central question of this project was an attempt to replicate, at scale, the results of published research. We selected 111 main effects and 22 interaction effects and, using a tool that helped construct appropriate LLM prompts, we tested each of the effects on two benchmarks: (1) the ability of the LLM to match the direction of effects reported in original publications, and (2) the ability of the LLM to match, using the exact same confidence intervals, the statistical results reported in the original research. Both of these tests relied on quantitative assessment of study results, comparing actual statistics from AI and human studies.

A second question concerned the generalization of results to messages, samples, measures, and research contexts not used in the original research. To explore this question, we conducted a qualitative assessment with a sample research result from a human study, using AI to run the same study numerous times with changes to the specifications of sampling, message stimuli, outcomes measures, and research contexts. All of the extensions that we tested might plausibly be assumed to be covered by the original study; that is, the new samples, stimuli, measures, and research contexts were not very different from the original human study that was replicated. But as our sample results show, and as many researchers now fear, even slight changes to study specifications can change the results. Our goal with these qualitative tests was to explore the possibilities for AI personas to contribute to generalization qualifications, and to do so more quickly and less expensively than could be expected in the natural course of human research calendars and budgets.

## Method

### *Sampling Studies for Replication with AI Participants*

To assess the replicability of marketing experiments using LLMs, we collected a sample of recent research that experimentally examined responses to different types of media messages. We systematically reviewed all articles in the *Journal of Marketing* published from January 2023 through May 2024, and that resulted in an initial corpus of 69 papers containing 210 unique studies. The *Journal of Marketing* was chosen for this initial test of AI replication accuracy because it frequently publishes tests of media message effectiveness (consistent with our interest in commercial and theoretical work about media psychology), and because journal policies usefully require detailed reporting about measures, sampling, and inclusion of actual visual and textual materials. All of that information is crucial for precise replication.

We then reviewed candidate studies to evaluate their suitability for AI-assisted replication. We applied the following inclusion criteria: (1) the study had to be a true experiment incorporating manipulated study conditions (not simply a survey that scored participant attitudes or beliefs or a study that compared correlations between individual difference variables and outcomes); (2) all original study materials (i.e., stimuli and measures) must have been provided by the authors or otherwise be publicly available; and (3) the study procedures or outcomes could not require physical actions or behavioral measures (e.g., eye-tracking, monitoring of subsequent real-life purchasing behavior). In essence, for this initial test, we selected experiments that were conducted in a social science laboratory or were conducted through online recruitment platforms like Mechanical Turk or Prolific. The selection process yielded 45 studies from 14 distinct research articles. All of the articles are listed in Table 1 in Appendix A.

Each study's experimental procedure, data collection, and analysis were then replicated using software designed to facilitate tests with AI personas and media. The software (described in detail in Appendix B [Methods Supplement]) was designed to test AI responses to different versions of multimodal media messages. The software allows researchers to input various media stimuli (images, videos, text, or any combination thereof), organize them into experimental conditions, specify participant characteristics, and define survey questions and scales. The program then generates responses from participants based on those parameters. For each study, a series of unique LLM instantiations were created in real time— one for each virtual persona— that exactly matched the sample distributions, characteristics, and context as specified in the published study report.

Each persona was then shown the exact text, image, and/or video stimulus, along with all other original study instructions. Separate AI instances for each virtual participant allowed instances to be aggregated, and differences to be tested statistically, in exactly the same way that the human subject statistical results were computed in the original research. This replication effort entailed generating a total of 19,447 virtual participants across the 45 studies.

***Strategy for Creating LLM-Powered Participants***

For each study, we generated an equivalent number of LLM-powered participants to match the sample size (N) of the original experiment. Each AI participant was generated as an individual instance of Anthropic's Claude Sonnet 3.5 (`claude-3-5-sonnet-20240620`), one of the most advanced LLMs available at the time of this research. We used a model temperature of 0.7, as this value was at the time of the study the industry standard (Argyle et al., 2023). The sample of participants was then imbued with specific persona characteristics mirroring the types, frequencies, and distributions of those reported in the original study. For example, a generated participant might be characterized as a "45-year-old woman with 20 years of managerial experience in the manufacturing industry."

The software then constructed a prompt that instructed the LLM to accomplish three tasks: 1) embody the assigned personas; 2) have each persona examine the presented stimuli; and 3) have each persona respond to the subsequent questions. The question wordings and response scales provided to the generated participants were taken directly from the original experiments. This approach allowed for flexibility in accommodating various question types and scale formats, ranging from open-ended queries (e.g., "What is the highest price you would be willing to pay for this product?") to Likert-style scales of varying points (e.g., 1 = very unlikely, 7 = very likely).

A crucial aspect of the method was ensuring that the LLM-powered participants remained unaware of the study goals or of the original study being replicated. The goal was to preclude the possibility that virtual personas would use study-specific training data to provide responses. Our primary interest was to assess whether an LLM could generate responses that resembled those from real human samples, and without direct access to the original material. To validate this approach, we conducted a pretest, using Claude 3.5 Sonnet, that queried the model's awareness of any of the published papers whose experiments we aimed to replicate. The model reported no prior knowledge of these studies, suggesting no direct threat to the integrity of our replication efforts. We also deliberately avoided mentioning any paper title, authors, or journal information when presenting the stimuli to the AI personas, helping to ensure that the responses were based on the given prompts rather than any prior knowledge.

A final test of model awareness for our project was conducted using a new technique called DE-COP (Duarte, et al., 2024) that is based on the premise that an LLM is highly likely to correctly identify verbatim excerpts from its training text. The method probes the LLM with multiple choice questions that ask "Which of the following responses is verbatim from [citation for article]"? The response options contain one verbatim string of text from the referenced paper, and three paraphrases of the verbatim text in which minor details were altered (e.g., "dinner" vs. "brunch"). The assumption is that if the model answered the multiple choice question correctly, then that particular study was more likely to have been included in the training set, whereas if the LLM got the question wrong, the model more likely did not have knowledge of the original study. For each test, we quizzed ten generic AI personas created with the selected LLM (Claude 3.5 Sonnet), asking each persona to pick one of the multiple choice options. Our tests of a subset ($n = 5$) of the sample studies articles suggested that the model did not have previous knowledge of the texts: in four cases the personas uniformly picked the same incorrect multiple choice option; in a fifth case 30% of the personas selected the correct option while the remaining 70% picked the same incorrect text snippet. These observations gave us additional confidence that the LLM had not memorized the study and was computing an answer using other data when completing the simulated study replications.

To create a dataset for each replicated study, each of the AI persona participants were linked to a unique API call to the LLM. After responding to the measures, each persona's answers were logged in a database. We then extracted these responses and employed R for statistical analysis, replicating the exact statistical procedures used by the original authors. This included various techniques such as ANOVA, linear models, and chi-square tests. A more detailed description of the process we used to build the panel of personas and generate the prompts that queried the LLM can be found in Appendix B (Methods Supplement).

***Assessing Replication of Original Human Participant Findings***

Our unit of analysis for replication was a predictive "finding" within a study (i.e., each main effect of an independent variable on a dependent variable, or interaction of multiple independent variables on a dependent variable). For each of the 45 studies, we compared our results to those reported in the original experiment, resulting in 133 replication observations.

Anderson and Maxwell (2016) defined several different goals for replication. The aims of our study aligned with their replication "Goal 1" — simply, "To infer the existence of a replication effect (p.3)." In turn, following their recommendation, we considered a previously significant finding successfully replicated if it matched the original study's result in direction and statistical significance ($p < .05$).) This criterion focused on reproducing the existence and direction of an effect, which Anderson and Maxwell argue is often sufficient, especially for unexpected or counterintuitive findings in psychological research. (We note that almost all of the previous tests using AI have focused on replication of direction only, and do not report replication of statistical significance.) Additionally, we attempted to replicate the null effects from studies that reported non-significant findings. We considered a non-significant finding successfully replicated if the LLM-based study also yielded a non-significant result ($p >= .05$). This approach acknowledges the importance of reproducing null effects, which may often be theoretically meaningful for theory validation and falsification (Meehl, 1978; Popper, 2005), for improving reliability (Ioannidis, 2005), and for demonstrating the relative robustness of past findings (e.g., Open Science Collaboration, 2015). It is important to note, however, that the replication test does not definitively prove the absence of an effect but rather demonstrates a failure to reject the null hypothesis in both the original study and replication.

Figure 1 illustrates a case study replicating an experiment about packaging design effects from Study 1a in Ton, et al. (2024). This study examined how complex versus simple packaging designs influenced consumer perceptions across four criteria: willingness to pay, inferences about ingredients, perceived product purity, and design attractiveness. For this replication, our test software generated 362 AI personas, matching the sample size and sample characteristics reported in the original study. The personas were presented with the same stimuli (complex or simple package designs) and then responded to measures identical to the human study. We then conducted parallel statistical analyses, comparing the direction and significance of effects observed in the original human data and the AI-generated data. The figure shows both the magnitude and statistical significance of experimental effects, allowing for a direct comparison between human and AI responses to these marketing stimuli.

## Results

### *Overall Replication Success Rate with Virtual Participants*

LLM replication tests successfully reproduced 76% of the original main effect findings (84 out of 111). This included both examination of the direction of effects and their level of statistical significance. When also accounting for interaction effects, the replication rate was 68% (90 out of 133). Notably, the replication rate for interaction effects alone was substantially lower at 27%. This difference in replication success for main and interaction effects parallels the differences for replication of human studies. Poorer replicability of interaction effects can be attributed to the higher statistical power needed for interaction tests, and to a greater potential influence of context-specific factors (Crede & Sotola, 2024). Additionally, as these statistical attributes of interaction effects lead to a higher proportion of non-significant findings compared to main effects, they may also lead to greater susceptibility to publication bias and questionable research practices (QRPs) that favor main effects (Ioannidis, 2008), particularly in the social sciences (Fanelli, 2010; Kerr, 1998).

### *Replication Success Rate By p Value Range*

Previous literature suggests several reasons why replication success may vary as a function of the *p* value of the originally observed effect. For instance, Ioannidis (2005) described how the probability of a finding being true depends partially on the level of statistical significance. And further, studies with *p* values nearer to .05 may actually be the result of biases in reporting, meaning they would in turn be less likely to replicate. Even without such biases, *p* values nearer to .05 may indicate greater likelihood that the original effect is marginal (Wasserstein & Lazar, 2016). Because of these biases, we examined the relationship between the *p* value of the original findings and the rate of successful replication. Figure 2 illustrates the relationship between the statistical significance (*p* values) of original findings and their replication success using LLMs.

The replication rate for findings with highly significant *p* values (below 0.001) was notably high, with 83% of the findings (54 out of 65) successfully replicated. This high replication rate indicates that LLMs are particularly effective in reproducing findings that are strongly supported by statistical evidence, reinforcing confidence in the ability of the models to capture robust phenomena. However, as the *p* values increase, indicating less statistically significant results, a general trend of declining replication success is observed. Specifically, for findings with *p* values in the range of 0.001 to 0.01, the replication rate drops to 53.3% (8 out of 15). In the 0.01 to 0.03 range, the replication rate slightly increases to 61.5% (8 out of 13) but then declines again to 55.6% (5 out of 9) for *p* values between 0.03 and 0.05. This pattern

Figure 1: Example Human Study and AI Replication: Packaging Design Effects

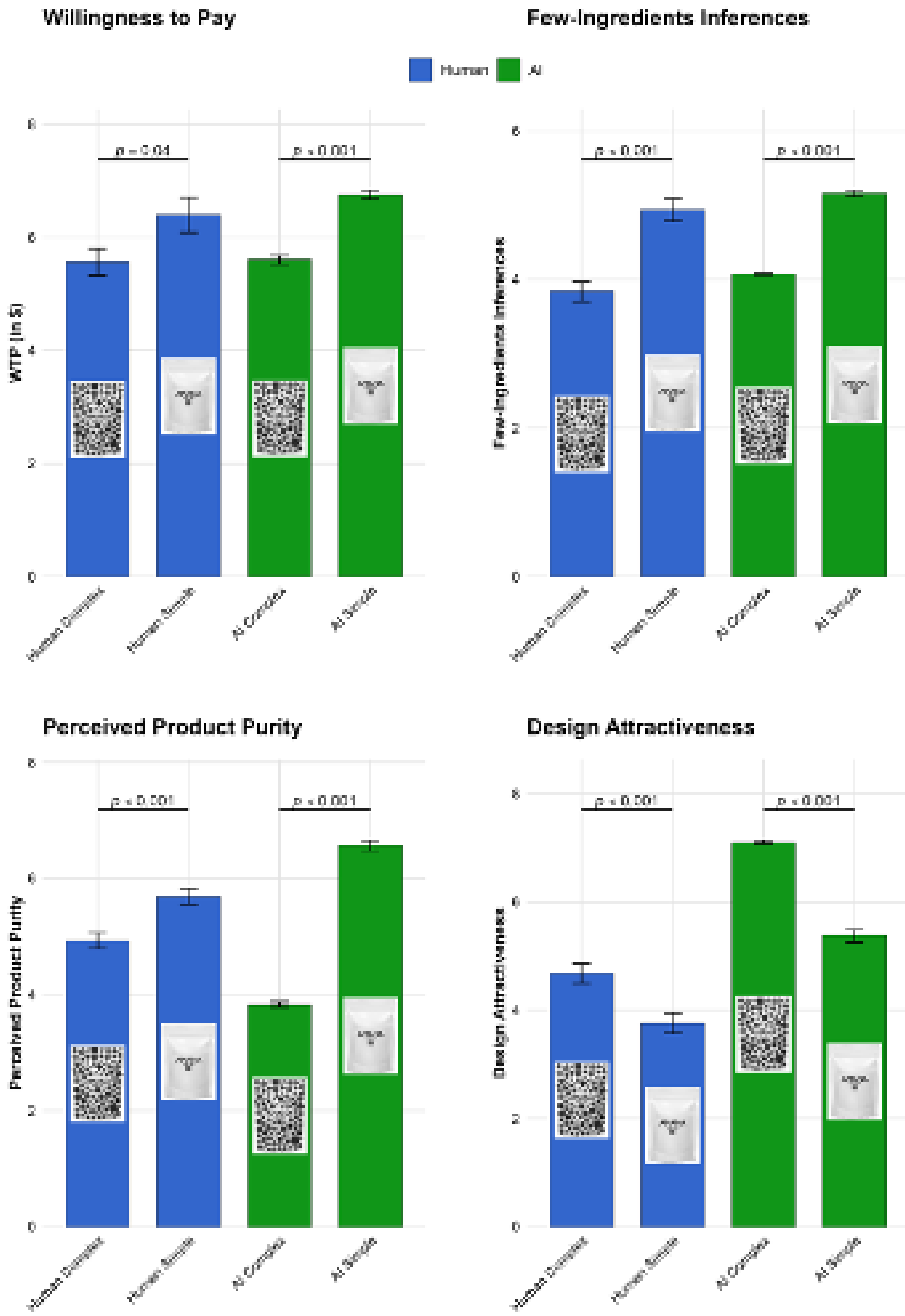

Notes: Replication of packaging design effects study (Ton, et al., 2024) comparing human and AI responses. AI persona responses showed effects in the same direction and level of statistical significance compared with the human study. Images in bars depict actual packaging stimuli used. Error bars represent ±1 SE.

Figure 2: Replication Success by *p* value Ra

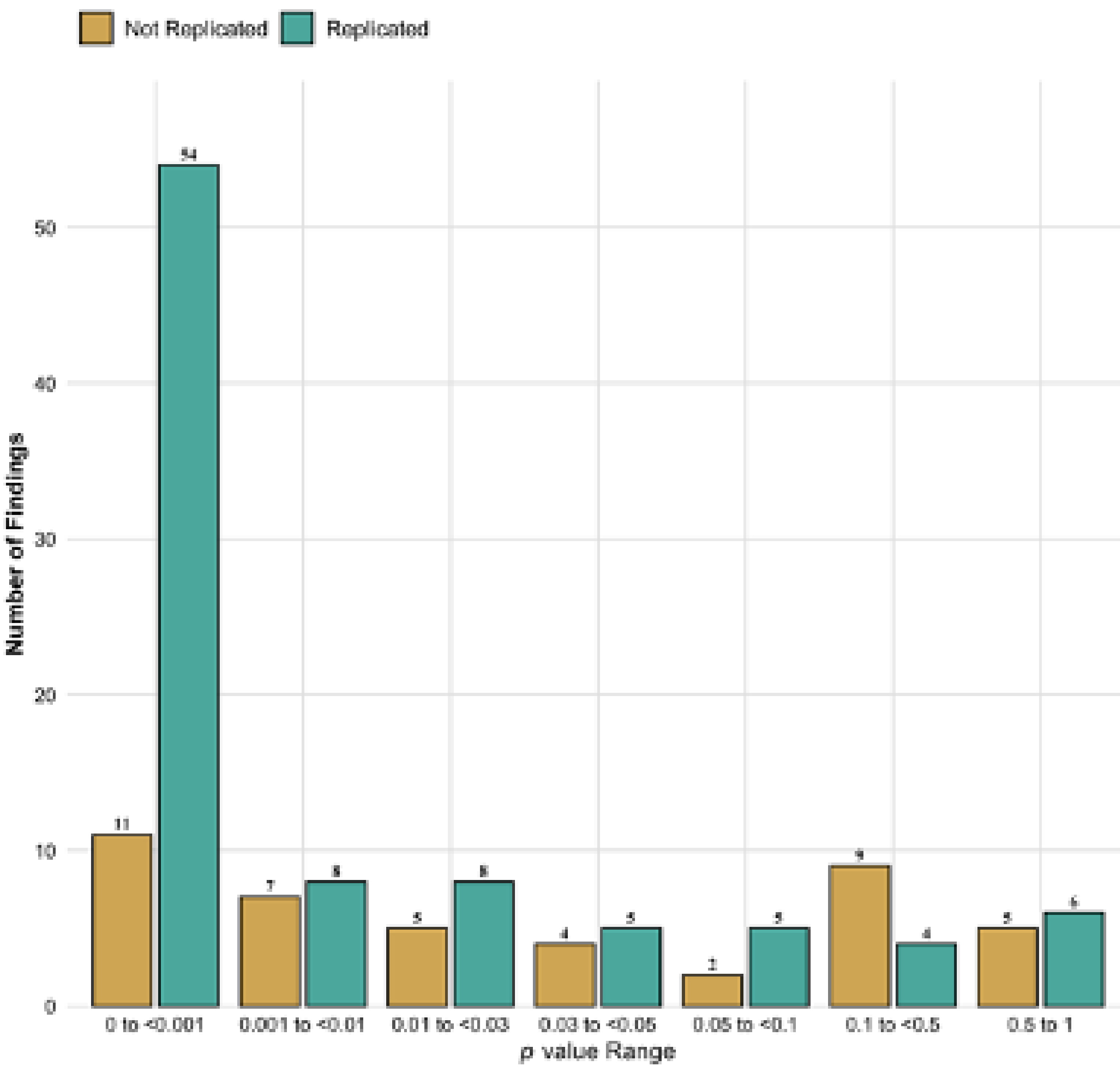

*Notes*: The graph shows the number of findings that were successfully replicated (green) or failed to replicate (purple) across different *p* value ranges taken from the original studies.

suggests that the strength of the original statistical evidence is a crucial factor in the successful replication of findings by LLMs. (See Table 2 in the Appendix A for a list of all the findings and associated p values.)

When examining findings with marginal significance (*p* values between 0.05 and 0.1), we observed a replication rate of 71% (5 out of 7). For findings in the 0.1 to 0.5 range, typically regarded as non-significant, the replication rate further decreased to 31% (4 out of 13).

Interestingly, for findings with $p$ values above 0.5, which generally suggest a lack of effect in the original studies, the replication rate increased to 55% (6 out of 11). While this is a small initial sample, this complex pattern highlights a potentially nuanced relationship between the statistical significance of original findings and their replicability by LLMs.

***Replication Success Rate by Effect Size***

Beyond statistical significance, it is generally the case that larger effects are more robust and easier to detect (Cohen, 2013; Ioannidis, 2005), and in turn also easier to replicate (Van Bavel et al., 2016). To test this with our sample of studies, we investigated the relationship between effect size of the original findings and the rate of successful replication.

For the findings that reported effect sizes (98 out of 103), we categorized the effect sizes into four ranges based upon the benchmarks set by Cohen (1988), and then we examined the replication rates for each category. The results are shown in Figure 3. There is a positive relationship between the magnitude of the original effect size and the likelihood of successful replication using LLM-powered participants. As the effect size increases, so does the rate of replication.

Figure 3: Replication Rate by Effect Size of Original Studies.

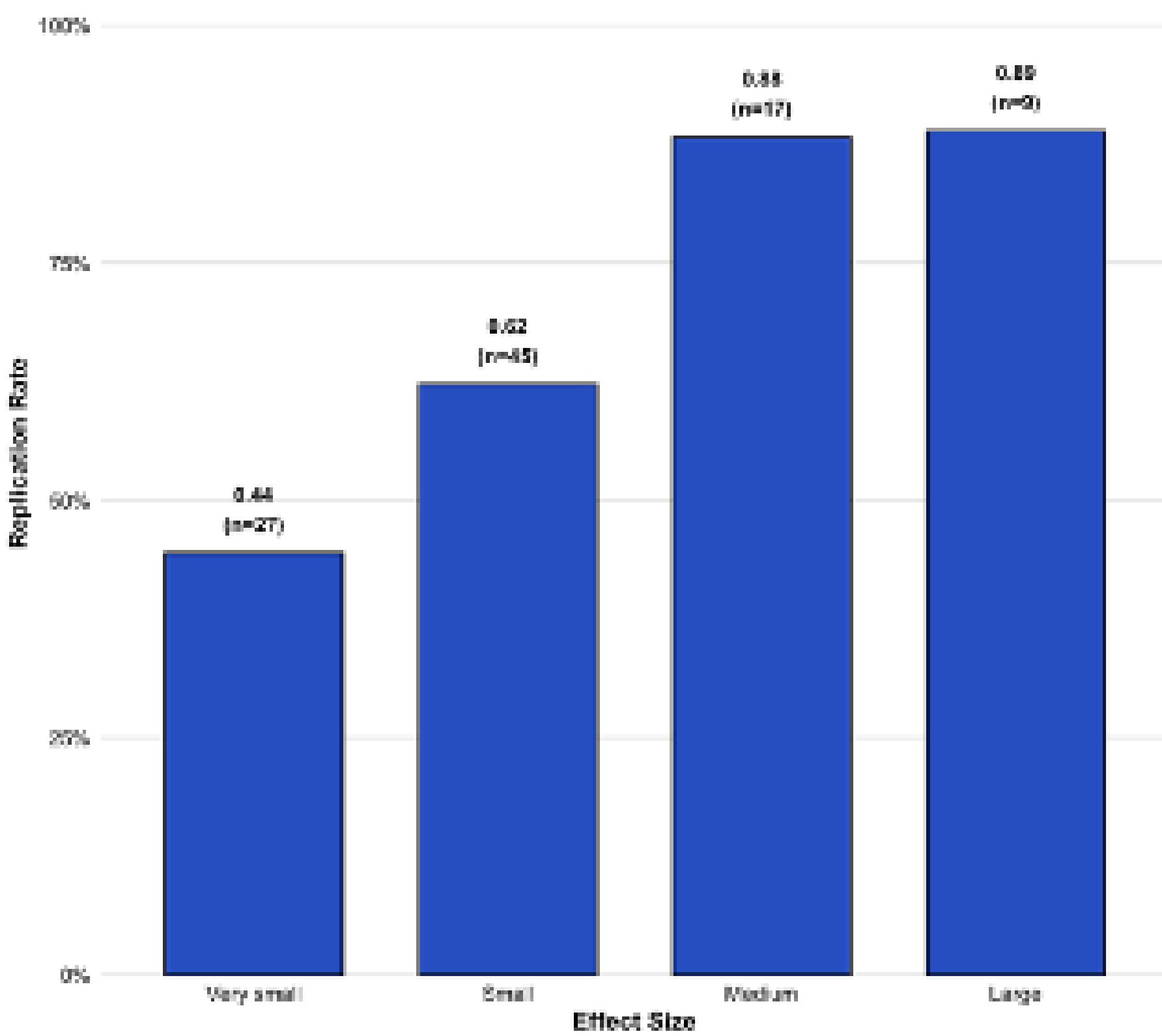

*Notes:* The effect size benchmarks for each of the four categories are suggested by Cohen (1988).

***Generalization of Results to Different Samples, Stimuli and Research Contexts***

The replication crisis was born out of concern about results remaining stable when replication tests shared identical features with the human studies. In marketing, however, the use of the results for both theory and practice depends substantially on findings that hold even when the application parameters are slightly changed. In other words, we hope that research findings will generalize to people, messages, and contexts that are different from the original study, even if they represent only minimal changes. We might ask, for example, if results of a study generalize from Gen Z to Millennials, from pricing studies of cameras to the pricing of kitchen appliances, or from participant samples recruited online to samples recruited in a shopping mall. Testing the limits of generalization is critical but rarely accomplished, and that poses a challenge to the accuracy of many if not most uses of research (Yarkoni, 2022). It is rare that applications of research share the exact same parameters as the studies used to generate the applied findings. Despite the importance, tests for generalizability are rare and they are difficult, due in large part to the practicalities of time, money, and availability of researchers. The speed and breadth of changes possible with LLM replications, however, offers opportunities to usefully explore generalizability constraints for theory development and practical application.

To demonstrate how LLMs might be used in generalization tests, we selected an example study for exploration (Maier, et al., 2024). There were 8 studies in the Maier, et al. publication, all designed to explore how and when the mention of university-industry collaborations in advertising made new products more attractive to consumers. We chose to explore different versions of Study 3, an online experiment that presented people with two different versions of a print advertisement for a new type of camera. In the original study, 102 participants, with a mean age of 37 and evenly split by gender, were provided with detailed information about two different cameras, including promotional text, pricing, and user reviews. In the first condition Camera A was said to be developed in collaboration with a university, while Camera B was said to be developed by the manufacturer alone. In the other condition, the co-developed label was switched to the Camera B. After reviewing the cameras, the participants were asked to answer the question “Which of the two products do you prefer?” as well as other evaluations of the scientific and technical features of the cameras. The main finding was that people significantly more often chose, and more positively evaluated, whichever product was co-developed with a university.

The initial LLM replication of the human study (part of the previously reported results) successfully reproduced this finding. Our task here was to experiment with different possible versions of the study, with each different version allowing for a test of reasonable extensions of the experiment that would show when and how the results might be used for refining theoretical understanding or practical application. We experimented with changes in four different categories: 1) different research contexts (e.g., online recruitment, live participation with experimenters in a laboratory); 2) different samples of people (e.g., generational differences, special audiences); 3) different measures of respondent attitudes (e.g. gauging product perceptions with different questions); and 4) different products that would appear in the advertising (e.g., cameras, appliances, digital services).

The variations in study features that might be useful to test are, of course, almost endless. Our goal with this exemplary study and generalizations from it was to show how the tests might be constructed with particular attention to versions of the study that might not replicate but that would be reasonable extensions for any project looking to use the findings of the original study. Here, we present 20 different changes to the original study, with commentary about whether the new AI studies suggest that generalization of the original findings is appropriate. A conclusion that the new studies replicated meant that the direction of the effects as well as the statistical significance of the findings were the same as those of the original human study, and the same as the initial AI replication that used the exact same study parameters. See Table 3 in Appendix A for a full summary of the tests reported below.

***Research Contexts***. In the original study, people participated in the research by viewing the ads on home computers via an interaction arranged through an online recruiting company. We tested the generalizability of the results to five AI studies with different arrangements for recruiting and viewing the ads. First, we instructed the LLM model that subjects were approached by a researcher while walking through a shopping mall. Second, we changed the recruiting context to the quadrangle in a large university. In a third test, we adjusted recruitment such that all participants were brought in through an advertisement seen on Reddit. Fourth, we changed the study scenario, in that the participants instead completed the experimental procedure in a campus laboratory space, with a one-way mirror on one of the room walls. In a fifth replication test, we changed the incentives given to participants: in the original study, participants were told that they would be entered in a lottery that would give them a chance to win the camera in the ads, and that because of that possibility, they should pay careful attention to the ad they would view. In a new version of the instructions, we eliminated all mention of the lottery and possible prize. In all five of these studies, the original findings from the original human study were replicated, though in the fifth case the aggregate effect of university co-development was primarily driven by one observations in one conditions (e.g., preferences for Camera B when co-developed [82%] rather than preferences for Camera A when co-developed [only 47%]). Nonetheless, these results, taken together, suggest the observed effect of university co-development on product preference may persist across varied recruitment and instructional contexts.

***Samples***. We next tested four different possible samples of respondents. First, we tested the original study with virtual personas that ranged in age from 18 to 22 years old (the original study used human subjects with an average of 37 years of age). Although the original study findings successfully replicated with AI personas that exactly matched the original study sample, they *did not* replicate with the "Gen Z" personas in this younger age range (rather, Camera B was generally preferred regardless of university co-development). Similarly, we did not replicate the original study's observed effect using AI personas that represented seniors aged 65 to 80, though, conversely, Camera A was generally preferred within this sample. Beyond age, we also considered how domain expertise may influence product perceptions. When we changed the sample to represent experts in photography (while retaining the mean age of the original human sample) the study with AI personas *did* replicate. Finally, we also explored how well these findings generalize to particular geographic regions. When recruiting a sample of AI personas from the rural Deep South of the United States, the original effect again failed to replicate, with Camera A being generally preferred in both conditions. These different outcomes due to changes in sampling suggest that the original results are sensitive to sampling details. And, in retrospect, it is not difficult to speculate about why these new AI samples might respond differently: younger people may not be impressed with the technical value of university collaboration; older people with less technical expertise may have been concerned about their ability to operate a complex camera; and expert photographers may have enthusiastically appreciated new innovations in their specialized hobby.

***Measures***. There are several other measures that could be used to ask about consumer responses to the value of university collaboration. We tested two. First, we changed the primary evaluative question to ask "Which of the two products would you prefer to give as a gift to your parents?" While the effect of co-development was still observed for Camera A, when Camera B was labeled as co-developed, preferences were much more evenly split, with only 49% of the AI personas selecting the co-developed product. When we changed the purchase decision question to one about buying the cameras for a son or daughter, the AI responses were uniformly in support of Camera B.

***Stimuli***. One of the most consequential threats to generalization concerns the sampling of stimuli to which the study might be applied. In this example study, the original research results would seemingly have application to a range of hi-tech camera products, as well as various other similar products. Exploring the breadth of applicability to similar products is crucial, but sampling stimuli in research is procedurally difficult because it requires building multiple versions of stimuli which is expensive and time consuming, especially if the research questions involve pictures or video. Further, testing additional stimuli presents constraints and threats to validity for both within-participant designs (e.g., fatigue, sensitization, order effects) and between-participant designs (i.e., more stimulus conditions equate to greater expense and time demand for recruitment, data collection, and compensation).

To examine whether the original findings were particular to the specific camera image stimuli used (which contained close-up photos of a hand presenting the small, portable, 360-degree camera), we first created new stimuli in which we substituted photos of comparable cameras. We also created a set of ads in which — given stimulus text referencing the portability and stabilization features of the devices — the images were replaced with photographs presenting skiers using comparable 360-degree cameras mounted to their helmets. These edited stimuli (see Appendix B [Methods Supplement] for renderings) allowed testing within the same general product category but with different yet plausible advertisement imagery. Running studies with each set of alternative stimuli, AI personas successfully reproduced the original study findings in both cases.

Second, as the initial study was not concerned with cameras in particular, we created two versions of ads (with and without university collaboration) for an assortment of different products whose technical function may conceivably be improved through such a collaboration (see renderings in Appendix B [Methods Supplement]). This included not simply replacing the product image itself, but in each case also creating novel promotional text and customer reviews corresponding with the new product category. Running new studies with these new products, we found that the original study results did not generalize equally across all product categories. For instance, the effect of university collaboration on consumer preferences was reproduced in judgments about tax service softwares and automatic espresso machines, but not for multiplayer video games about space conquest, washing machines, electronic muscle stimulators, all-natural skin balms, or neuro-linguistic programming apps. This demonstrates the need to carefully evaluate how conceptually distant a product can be before results change and conclusions potentially no longer hold.

The AI findings obtained with different versions of the original experiment suggest that the concern about generalization is well founded. Not only must the exact results of a study be replicated to build confidence in the research, but defining the boundaries of generalization will be critical before we can know how to use the results for theory and application. A significant problem, one we will discuss in the next section, is that these boundary tests, even more so than exact replications, are *unlikely* to be conducted, not because they aren't needed, but because such studies are time consuming, expensive, logistically difficult, numerous with respect to possibilities for extension, and currently not considered good evidence of professional development.

## Discussion

Our study suggests that empirical studies about media message processing can use AI virtual personas to replicate existing research, and especially the main effects proposed in such studies. The level of replication, while not perfectly matched with identical human studies, is at a level of success (76%) that offers promise to help solve an important challenge to current social research. We note that this replication rate compares quite favorably with known rates of replication using human subjects to replicate human studies. We conclude that AI personas do at least as well as human replications and perhaps better.

Not all original study results replicated similarly, however, with implications for the capabilities and limitations of LLMs for consumer behavior research. First, the high reliability of LLMs in replicating strongly significant findings suggests a primary value for *confirming* robust effects. Second, the observed decline in replication success as $p$ values increase emphasizes the critical role of the strength of the original evidence. Stronger original evidence is more likely to be successfully replicated by LLMs, which is an important consideration for researchers that use these models. Third, the mixed performance of LLMs on findings with marginal or non-significant $p$ values raises concerns about AI sensitivity to subtle effects. This variability suggests that LLMs may be prone to both false positives and false negatives, indicating a potential risk when using them to detect or confirm less pronounced effects. Last, the balanced replication outcomes for *p* values above 0.5 shows that while LLMs may sometimes accurately identify the absence of an effect, they also risk introducing spurious findings. This dual possibility underscores the need for caution when interpreting LLM-based replications, particularly in cases where the original findings suggest a null effect.

The value of AI replication could come in several forms. Most obviously, the replications, when they agree with identical human subject studies, can provide evidence that many existing results can be used with confidence, and we would say more confidence than in the absence of any replication, either human or AI. When the AI replications do not work, this might identify which results should be a priority for replications with human subjects so that differences can be adjudicated. It's unlikely that we will ever have the capacity to replicate all of the research we'd like to use, so knowing which studies represent the greatest risk to error in theory development or practical application would be useful. Not only do the demands of time and money for individual researchers often preclude replication, but replication is still considered questionable as a professional contribution for academic advancement. Beyond individual scholars, replication projects organized by groups of researchers (e.g., Baumeister, et al., 2023), might similarly use the AI results to prioritize field-wise research agendas and to review progress in the field more generally.

Replication possibilities with AI represent a strong promise for media and marketing research, similar to other uses that may allow AI to impersonate humans in simulated media environments, create experimental media stimuli, or generate plausible human behavior that can be placed within media presentations. All of these applications depend to varying degrees on AI being able to offer insight into how human intelligence works, which is the promise that garners the most discussion for both scientific advancement as well as apocalyptic worries. Perhaps equally as influential in the near term, however, is the ability of AI to just make research faster and cheaper – possibly by several orders of magnitude. We mention possible positive changes in two different media message research contexts, at opposite ends of an applied-theory continuum.

***Applications***. Much of applied research is interested in tests of alternatives for persuasive messages, often in the context of message design exercises that proceed quickly and without significant resources. For example, researchers might want to pretest versions of a health PSA designed to change behavior, TV advertisements designed to solidify brand perceptions, or social media posts seeking to promote clicks. In any of these applied settings, new studies using AI personas could be conducted in minutes, maybe even during the one-hour meeting where designers propose, test, and select finalist messages. And as our example results about research generalization suggest, multiple versions of similar tests might be conducted in that same meeting to explore subtle differences in language use, differences among children, adolescents and adults, or differences in the US, EU, and Asia. Designers could also quickly test the effects of placing messages in different contexts (e.g., different social media platforms). Responses could be assessed within minutes, and the results plugged into design discussions in close to real time.

***Theory***. Theory in media psychology could be similarly advantaged by speed and cost. Theory development that requires the sequencing of questions could proceed more quickly (e.g., depending on the results of a first study, we could move to one of two second questions). The significant advantage in theory building with AI personas is the ability to quickly and purposefully alternate between induction and deduction. That is, sequencing questions about the description of what people are doing as separate from explanations of why and how those actions might happen. Currently, theory in media psychology privileges deduction; that is, researchers are encouraged to deduce a hypothesis from a theory and then proceed to an empirical test. Often, however, deductions are premature because we do not yet have adequate descriptions of the media behaviors and perceptions (i.e., inductive input) that new theories should be about. AI studies could accelerate transitions in the inductive-deductive cycle.

The ability to move quickly in research also connects well with the extremely fast pace at which media technologies and their uses are changing. As media psychologists, we volunteer to link theories with characteristics of media that are currently dominant; for example, if smartphone use is fragmented into extremely short and unrelated segments, we might emphasize theories about task switching to address the effects of that media trend. But the most interesting features are constantly changing, often within the time it takes to conduct new studies. Currently, much of media psychology theory is based on tests with content and technology of decades past, and especially with television. New technologies dramatically shift what theories should be developed to make certain we are studying the most essential features of the stimuli that are most responsible for current scientific interests and consumer practices.

***Accuracy and Bias in AI Replications***

Any excitement about AI personas depends crucially on accuracy. There is not perfect agreement between AI and human studies, and while the absolute level of agreement is high (76%), there is still uncertainty about errors, both within the matching and divergent results. Here, we review three study features that determined the level of AI-Human matching in our project, and other general issues that will be critical to understand as AI persona research develops

*Interaction vs. main effects.* Main effects were substantially easier to replicate with AI personas (76%) than were interaction effects (27%), suggesting that interaction effects may be particularly prone to false positives or inflated effect sizes. This is a pattern that has been identified in human subject replications of social research, and the differences between main and interaction effect results in those studies are similar to our finding. Crede and Sotola (2024) examined 244 tests of interaction effects from leading organizational science journals and estimated an overall replicability rate of only 37% using z-curve analysis. They found that over half of reported $p$ values for interactions fell between .01 and .05, far higher than expected for well-powered studies with true effects. Similarly, Altmejd, et al. (2019) suggested that the study attribute most predictive of poor replicability is whether central tests describe interactions between variables or (single-variable) main effects. In their review of organizational behavior studies, only eight of 41 interaction effect studies replicated, while 48 of the 90 other studies did.

One explanation for the lower replicability of interaction effects may be that holding sample size constant, interactions will have lower statistical power. In addition, they also represent more complex conceptual expectations, perhaps so much so that they do not reflect reality. Sherman and Pashler (2019) analyzed five large-scale datasets—including thousands of participants and hundreds of demographic and psychological predictors—and found that interaction effects are generally small, infrequent, and likely due to sampling error. Thus, while virtual AI participants may not permit substantive replication of interaction effects reported in the existing literature, such an approach appears no worse than what can be gleaned through replication with human subjects and may actually permit a faster, less expensive route towards assessing the relative validity of previously reported moderation findings.

*Ground Truth.* There are several possible explanations for mismatches between human subject studies and AI persona replications. In our analysis, about 1 in 4 of the statistically significant main effects reviewed resulted in no significant differences when using AI personas. Notably, there were no cases where the two methods disagreed in the *direction* of effects; differences were only about whether they could be viewed as statistically significant.

Reconciling the differences is not (yet) straightforward. There are two different arguments for determining which results, those found with human participants or those from studies with AI personas, should represent the most accurate characterization of an effect when they are unaligned. So far in the literature, ground truth is mostly defined as a result based on human data. We know, however, that there are multiple critiques of human studies that limit confidence in the conclusions from those experiments. The critiques prominently include biases associated with gender, race, age, and cultural context, and as well as growing concern about the validity of data from online data collection. Since the LLM models are trained on information that also includes those biases, it is possible (and some argue likely) that biases will be transferred to the AI models. With AI replication, however, the biases might be made less influential because studies could be focused so as to emphasize samples of different (and hard to acquire) demographic groups or to change the methods by which data are collected. With respect to sampling, more inclusive AI samples might even flip the ground truth assumptions, making AI a more deserving gold standard for representativeness.

*Stimulus Sampling.* The generalizability crisis underscores the danger of making broad claims based on narrow samples of people, stimuli and research procedures. The sampling of stimuli is particularly troublesome in media psychology (Reeves et al., 2016). While there is substantial investment in sampling human subjects in research about responses to media, there is a significant *underinvestment* in the sampling of media stimuli themselves. This imbalance poses a serious threat to the external validity and to the overall usefulness of research (Cummings & Reeves, 2022). Media research most often relies on single examples of media, and even when there are several examples they are often poor representations of the natural variance in a category of stimuli. This practice overlooks extensive variability inherent in media content, which can lead to erroneous conclusions that undermine the validity of research.

Westfall et al. (2015) emphasized that research must resample stimuli in replication attempts to increase confidence in original findings. Relying solely on participant replication without considering stimulus variation, they argued, can introduce unintended variables that alter results. By not adequately sampling media stimuli, researchers risk committing Type I errors where they falsely identify effects that do not generalize beyond the specific stimuli used. This contributes to the replication crisis, as subsequent studies using different stimuli may fail to reproduce the original findings, but primarily because they use different stimuli rather than different participants. AI tools may help substantially with this problem because they can more easily produce new samples of media, a task that would require significant resources and time if done equally rigorously by and with humans.

***Challenges with AI and Social Research***

One of the greatest challenges to the evaluation of AI applications is the inability of researchers to examine training data from proprietary LLMs. Consequently, it is difficult to understand how biases of the internet and other training data might affect the accuracy of models. AI represents one of the first major technology developments that has flipped the progression of research from university labs to technology companies. For AI, it is the latter who now own access to the models and other labs who are trying to understand how the models work. There are important new calls for development of open LLMs that will allow appropriate experimentation and knowledge about training data and those will be critical for developing social research applications of AI (Li, 2023).

Even working within the constraints of current commercial models, there are parameters of the models that can be manipulated by researchers to increase the value of the technology. Two important features are the specification of prompts that translate variables and measures into information on which LLMs can operate, and the temperature settings that models provide that control the variance (or determinism) of models. In this research, we used a temperature of 0.7. This resulted in some instances where the LLM's responses showed less variance, as measured by standard deviations, compared to human responses for the same findings. Higher temperatures may lead to more variance in LLM responses, which could mitigate the tendency for LLMs to produce homogenous results.

We mention one additional limitation of this project that is generally applicable to all current uses of AI in social research. We do not claim, nor do any of the AI attempts to replicate human studies, that the LLM models are achieving accurate results because they are thinking like a human. We do, however, believe that the AI models are usefully predicting the results of human studies about thinking, but not via development of any advanced formula that explains what is contained in the black box of human cognition. Consequently, an important limitation of this and other similar projects is that they *do not provide full explanations* of how humans think, although researchers do hold out the prospect that future AGI projects can move in this direction.

We do, however, believe there is significant value in prediction, even absent complete explanations. Arguments about prediction vs. explanation have always been part discussions about computation. For example, linguists thought it necessary to understand how language works in humans (i.e., explanations of the "black box") before applying that knowledge to technology like speech recognition. To make predictions, however, that turned out not to be the case, as computer scientists were able to make the predictions necessary to recognize speech, make medical diagnoses, and discover new drugs mostly by brute force computation (Norvig, 2017). Explanations ("how does the black box work?) are still a goal, but predictions have value right now. In marketing science, we believe that as long as the predictions are accurate, there is significant value in both applications of results, as well as a supportive role in the advancement of theory, largely by providing confidence in research results that can be the building blocks of theory construction.

***Conclusion***

We believe the efficiency rationale for pursuit of AI personas is compelling. The exercise in this project simulated 133 empirical tests that were originally conducted with 19,447 human subjects by 47 different researchers and published in 14 articles with publication timelines that take months and more often years to complete. The money spent is likely in the tens or hundreds of thousands of dollars (and millions counting salaries), and the time spent by 47 researchers would be in the months and years. Importantly, adequate human replication designs may be even more costly as recent work suggests replication studies require samples 16 times that of the original study to achieve the 80% of the power required for detecting the original effect with a $p$ value of .05 (van Zwet & Goodman, 2022). Our replication studies, consisting of nearly 20,000 AI personas, were conducted with tens of dollars and, once the prompting techniques were designed, they were completed in only a few hours. That is a huge resource and time advantage for both theory and application, certainly enough to warrant continued evaluation of AI personas.

## Appendix A: Summary Tables.

Table 1: Papers from the Journal of Marketing Chosen for Replication.

| Paper | Citation |
|---|---|
| 1 | Garvey, A. M., Kim, T., & Duhachek, A. (2023). Bad news? Send an AI. Good news? Send a human. *Journal of Marketing, 87*(1), 10-25. |
| 2 | Chen, J., Xu, A. J., Rodas, M. A., & Liu, X. (2023). Order matters: Rating service professionals first reduces tipping amount. *Journal of Marketing*, *87*(1), 81-96. |
| 3 | Wang, P. X., Wang, Y., & Jiang, Y. (2023). Gift or donation? Increase the effectiveness of charitable solicitation through framing charitable giving as a gift. *Journal of Marketing, 87*(1), 133-147. |
| 4 | Herhausen, D., Grewal, L., Cummings, K. H., Roggeveen, A. L., Villarroel Ordenes, F., & Grewal, D. (2023). Complaint de-escalation strategies on social media. *Journal of Marketing*, *87*(2), 210-231. |
| 5 | Biraglia, A., Fuchs, C., Maira, E., & Puntoni, S. (2023). When and why consumers react negatively to brand acquisitions: A values authenticity account. *Journal of Marketing*, *87*(4), 601-617. |
| 6 | Esterzon, E., Lemmens, A., & Van den Bergh, B. (2023). Enhancing Donor Agency to Improve Charitable Giving: Strategies and Heterogeneity. *Journal of Marketing*, *87*(4), 636-655. |
| 7 | Affonso, F. M., & Janiszewski, C. (2023). Marketing by design: The influence of perceptual structure on brand performance. *Journal of Marketing*, *87*(5), 736-754. |
| 8 | Berger, J., Moe, W. W., & Schweidel, D. A. (2023). What holds attention? Linguistic drivers of engagement. *Journal of Marketing, 87*(5), 793-809. |
| 9 | Costello, J. P., Walker, J., & Reczek, R. W. (2023). "choozing" the best spelling: Consumer response to unconventionally spelled brand names. *Journal of Marketing*, *87*(6), 889-905. |
| 10 | Maier, L., Schreier, M., Baccarella, C. V., & Voigt, K. I. (2024). University Knowledge Inside: How and When University–Industry Collaborations Make New Products More Attractive to Consumers. *Journal of Marketing*, *88*(2), 1-20. |
| 11 | Villanova, D., & Matherly, T. (2024). For shame! Socially unacceptable brand mentions on social media motivate consumer disengagement. *Journal of Marketing, 88*(2), 61-78. |
| 12 | Ton, L. A. N., Smith, R. K., & Sevilla, J. (2024). Symbolically simple: How simple packaging design influences willingness to pay for consumable products. *Journal of Marketing*, *88*(2), 121-140. |
| 13 | Tari, A., & Trudel, R. (2024). Affording disposal control: the effect of circular take-back programs on psychological ownership and valuation. *Journal of Marketing*, *88*(3), 110-126. |
| 14 | Mani, S., Astvansh, V., & Antia, K. D. (2024). Buyer–Supplier Relationship Dynamics in Buyers' Bankruptcy Survival. *Journal of Marketing, 88*(3), 127-144. |

*Notes:* The selected papers are from the Jan 2023-May 2024 issues of Journal of Marketing.

Table 2: Full Results of 133 Experimental Finding Replications.

| Paper ID | Study | N | Human *p* value | AI *p* value | Dependent Variable Finding | Replicated Finding? |
|---|---|---|---|---|---|---|
| 1 | 2 | 698 | < .001 | < .001 | Offer acceptance likelihood - main effect of offer type | Yes |
| 1 | 2 | 698 | 0.8 | < .001 | Offer acceptance likelihood - main effect of agent type | No |
| 1 | 2 | 698 | 0.001 | < .001 | Interaction - offer type x agent type | No |
| 1 | 3b | 400 | < .001 | < .001 | Offer satisfaction - main effect of offer type | Yes |
| 1 | 3b | 400 | 0.87 | 0.534 | Offer satisfaction - main effect of agent | Yes |
| 1 | 3b | 400 | 0.002 | 0.468 | Interaction - offer type x agent type | No |
| 2 | 2 | 146 | 0.001 | < .001 | Tip amount as a function of order | Yes |
| 2 | 2 | 146 | 0.05 | < .001 | Tip amount as a function of service quality | Yes |
| 2 | 2 | 146 | 0.98 | 0.075 | Interaction - order x service quality | Yes |
| 2 | 2 | 146 | < .001 | < .001 | Rating scores - main effect of service quality | Yes |
| 2 | 2 | 146 | 0.09 | < .001 | Rating scores - main effect of order | No |
| 2 | 2 | 146 | 0.1 | 0.349 | Interaction - order x service quality | Yes |
| 2 | 3 | 709 | < .001 | < .001 | Tip amount - main effect of payor | Yes |
| 2 | 3 | 709 | 0.2 | < .001 | Tip amount - main effect of order | No |
| 2 | 3 | 709 | 0.02 | 0.071 | Interaction - payor x order | No |
| 2 | 3 | 709 | 0.21 | NA[a] | Influence of order on rating scores | Yes |
| 2 | 4 | 349 | 0.04 | 0.001 | Tip amount | Yes |
| 2 | E | 120 | 0.04 | < .001 | Tip amount | Yes |
| 3 | 1 | 456 | 0.033 | 0.351 | Intention to contribute | No |
| 3 | 1 | 456 | 0.72 | 0.116 | Expected reciprocity | Yes |
| 3 | 4 | 299 | < .001 | NA[b] | Perception of gift vs. donation | No |
| 4 | 3a | 850 | < .001 | < .001 | Gratitude - main effect of arousal | Yes |
| 4 | 3a | 850 | < .001 | < .001 | Gratitude - main effect of active listening | Yes |
| 4 | 3a | 850 | 0.096 | < .001 | Interaction - arousal x active listening | Yes |
| 4 | 3b | 851 | < .001 | < .001 | Gratitude - main effect of arousal | Yes |
| 4 | 3b | 851 | < .001 | < .001 | Gratitude - main effect of empathy | Yes |
| 4 | 3b | 851 | 0.526 | < .001 | Interaction - arousal x empathy | No |
| 5 | 1 | 404 | < .001 | < .001 | Product choice | Yes |
| 5 | 1 | 404 | < .001 | < .001 | Brand values authenticity | No |
| 5 | 3a | 424 | < .001 | < .001 | Purchase intent - acquired vs. not acquired | Yes |
| 5 | 3a | 424 | 0.659 | 0.001 | Purchase intent - company size | No |
| 5 | 3b | 400 | 0.005 | 0.484 | Purchase intent - acquired vs. not acquired | No |
| 5 | 3b | 400 | 0.661 | 0.362 | Purchase intent - company size | Yes |
| 5 | 4 | 424 | < .01 | < .01 | Purchase intent | Yes |
| 5 | 4 | 424 | < .01 | < .001 | Brand values authenticity | Yes |
| 5 | 5 | 363 | 0.003 | 0.465 | Purchase intent | No |
| 5 | 5 | 363 | < .001 | < .001 | Brand values authenticity | Yes |
| 5 | 5 | 363 | 0.06 | < .001 | Product quality | No |
| 5 | 6 | 360 | < .001 | 0.184 | Purchase intent | No |
| 5 | 6 | 360 | < .001 | < .001 | Brand values authenticity | Yes |
| 5 | 6 | 360 | < .001 | < .001 | Product quality | Yes |
| 5 | 6 | 360 | < .001 | < .001 | Underdog status | Yes |
| 5 | 7 | 361 | 0.002 | < .001 | Purchase intent | Yes |
| 5 | 8 | 504 | < .001 | 0.5562 | Purchase intent - main effect of brand acquisition factor | No |
| 5 | 8 | 504 | 0.21 | 0.1025 | Purchase intent - main effect of growth orientation | Yes |
| 5 | 8 | 504 | 0.028 | 0.1611 | Interaction effect - brand acquisition factor x growth orientation | No |
| 5 | 9 | 633 | < .001 | 0.103 | Purchase intent - main effect of acquisition | No |
| 5 | 9 | 633 | 0.889 | 0.602 | Purchase intent - brand age | Yes |
| 5 | 9 | 633 | 0.039 | 0.255 | Interaction effect - acquisition x brand age | No |
| 6 | 1a | 304 | < .001 | < .001 | Sense of agency | Yes |
| 6 | 1b | 304 | < .001 | < .01 | Sense of agency | Yes |
| 6 | 2 | 401 | 0.1 | 0.1 | Interaction - targeting-via-options x targeting-via-amounts | Yes |
| 6 | 2 | 401 | 0.1 | 0.1 | Amount donated - main effect of targeting-via-options | Yes |
| 6 | 2 | 401 | < .001 | 0.24 | Amount donated - main effect of targeting-via-amounts | No |
| 7 | 4 | 800 | < .001 | 0.341 | Interaction - perceptual structure x hedonic positioning | No |
| 8 | 2 | 278 | < .001 | 0.002 | Sustaining attention | Yes |
| 8 | 2 | 278 | 0.009 | < .001 | Uncertainty | Yes |
| 8 | 2 | 278 | 0.001 | < .001 | Emotional language | Yes |
| 9 | 3b | 400 | 0.026 | < .001 | Download intentions | Yes |
| 9 | 3b | 400 | < .001 | < .001 | Persuasion tactic | Yes |
| 9 | 3b | 400 | < .001 | < .001 | Sincerity | Yes |
| 9 | 3b | 400 | 0.47 | 0.022 | Coolness | No |
| 9 | 3b | 400 | 0.024 | < .001 | Competency | Yes |
| 9 | 3b | 400 | < .001 | < .001 | Novelty | Yes |
| 9 | 3b | 400 | < .001 | < .001 | Processing fluency | Yes |
| 9 | 5 | 398 | < .001 | 0.004 | Purchase intentions - main effect of brand name spelling | Yes |
| 9 | 5 | 398 | < .001 | < .001 | Purchase intentions - main effect of naming motives | Yes |
| 9 | 5 | 398 | < .001 | 0.797 | Interaction - brand name spelling x naming motives | No |
| 9 | 6 | 406 | 0.06 | 0.651 | Purchase intent - main effect of brand name spelling | Yes |
| 9 | 6 | 406 | 0.009 | < .001 | Purchase intent - main effect of memorability | Yes |
| 9 | 6 | 406 | < .001 | 0.002 | Interaction - brand name spelling x memorability | Yes |
| 10 | 3 | 102 | 0.029 | < .001 | Product choice | Yes |
| 10 | 3 | 102 | < .001 | < .001 | Scientific legitimacy | Yes |
| 10 | 4 | 200 | < .001 | 0.748 | WTP | No |
| 10 | 5 | 785 | 0.008 | 0.029 | WTP - main effect of codevelopment | Yes |
| 10 | 5 | 785 | < .001 | 0.049 | WTP - main effect of product type | Yes |
| 10 | 5 | 785 | 0.002 | 0.132 | Interaction - codevelopment x product type | No |
| 10 | 6 | 444 | 0.045 | 0.299 | Purchase intent - main effect of codevelopment | No |
| 10 | 6 | 444 | 0.94 | < .001 | Purchase intent - main effect of firm type | No |
| 10 | 6 | 444 | 0.003 | 0.503 | Interaction - codevelopment x firm type | No |
| 11 | 3 | 295 | < .001 | < .001 | Disengagement intentions | Yes |
| 11 | 3 | 295 | 0.018 | < .001 | Vicarious shame | Yes |
| 11 | 3 | 295 | 0.5 | < .001 | Brand attitudes | No |
| 12 | 1a | 362 | 0.03 | < .001 | WTP | Yes |
| 12 | 1a | 362 | < .001 | < .001 | Few-ingredients inferences | Yes |
| 12 | 1a | 362 | < .001 | < .001 | Perceived product purity | Yes |
| 12 | 1a | 362 | < .001 | < .001 | Design attractiveness | Yes |
| 12 | 1a | 362 | 0.308 | < .001 | Design familiarity | No |
| 12 | 1b | 346 | < .001 | < .001 | WTP | Yes |
| 12 | 1b | 346 | < .001 | < .001 | Few-ingredients inferences | Yes |
| 12 | 1b | 346 | < .001 | < .001 | Perceived product purity | Yes |
| 12 | 1b | 346 | 0.376 | < .001 | Design attractiveness | No |
| 12 | 1b | 346 | 0.016 | < .001 | Design familiarity | Yes |
| 12 | 2 | 358 | 0.004 | < .001 | WTP | Yes |
| 12 | 3 | 798 | 0.21 | 0.046 | WTP - main effect of packaging design | No |
| 12 | 3 | 798 | 0.02 | < .001 | WTP - main effect of ingredient list | Yes |
| 12 | 3 | 798 | 0.038 | 0.152 | Interaction - packaging design x Ingredient list | No |
| 12 | 3 | 798 | < .001 | < .001 | Few-ingredients - main effect of packaging design | Yes |
| 12 | 3 | 798 | < .001 | < .001 | Few-ingredients - main effect of ingredient list | Yes |
| 12 | 3 | 798 | 0.03 | 0.564 | Interaction - packaging design x ingredient list | No |
| 12 | 3 | 798 | < .001 | < .001 | Product purity - main effect of packaging design | Yes |
| 12 | 3 | 798 | < .001 | < .001 | Product purity - main effect of ingredient list | Yes |
| 12 | 3 | 798 | 0.02 | < .001 | Interaction - packaging design x ingredient list | No |
| 12 | 5 | 742 | 0.565 | < .001 | WTP - main effect of packaging design | No |
| 12 | 5 | 742 | 0.263 | < .001 | WTP - main effect of consumption goal | No |
| 12 | 5 | 742 | 0.001 | < .001 | Interaction - packaging design x consumption goal | No |
| 12 | 5 | 742 | < .001 | < .001 | Few-ingredients inferences - main effect of packaging design | Yes |
| 12 | 5 | 742 | 0.004 | < .001 | Few-ingredients inferences - main effect of consumption goal | Yes |
| 12 | 5 | 742 | 0.361 | < .001 | Interaction - packaging design x consumption goal | No |
| 12 | 5 | 742 | < .001 | < .001 | Perceived product purity - main effect of packaging design | Yes |
| 12 | 5 | 742 | < .001 | < .001 | Perceived product purity - main effect of consumption goal | Yes |
| 12 | 5 | 742 | 0.33 | < .001 | Interaction - packaging design x consumption goal | No |
| 13 | 1b | 300 | < .001 | NA[a] | WTP | Yes |
| 13 | 2a | 300 | 0.047 | 0.004 | WTP | Yes |
| 13 | 2a | 300 | 0.007 | 0.39 | Psychological ownership | No |
| 13 | 2a | 300 | < .001 | < .001 | Quality | Yes |
| 13 | 2a | 300 | 0.002 | < .001 | Convenience | Yes |
| 13 | 2a | 300 | < .001 | < .001 | Environmental responsibility | Yes |
| 13 | 2a | 300 | 0.034 | < .001 | Perceived cost to the company | Yes |
| 13 | 2a | 300 | 0.255 | 0.441 | Novelty | Yes |
| 13 | 2a | 300 | 0.01 | 0.318 | Wastefulness | No |
| 13 | 2b | 300 | 0.009 | < .001 | WTP | Yes |
| 13 | 2b | 301 | < .001 | < .001 | Control | Yes |
| 13 | 2b | 301 | < .001 | < .001 | Psychological ownership | Yes |
| 13 | 2c | 300 | 0.016 | 0.109 | WTP | No |
| 13 | 2c | 300 | < .001 | < .001 | Disposal control | Yes |
| 13 | 2c | 300 | < .001 | < .001 | Psychological ownership | Yes |
| 13 | 3 | 807 | 0.855 | 0.544 | WTP - main effect of circular program | Yes |
| 13 | 3 | 807 | 0.41 | 0.55 | WTP - main effect of purchase target | Yes |
| 13 | 3 | 807 | 0.021 | < .001 | Interaction - Circular program x purchase target | Yes |
| 13 | 4a | 300 | < .001 | < .001 | WTP | Yes |
| 13 | 4b | 444 | < .001 | < .001 | WTP | Yes |
| 14 | 2 | 221 | < .001 | < .001 | Likelihood of approving debt structure | Yes |

[a]Indicates that the AI study did not produce enough variance for a meaningful *p* value. This occurred in two cases.
[b]Indicates that the study's manipulation check did not pass. This occurred in one case.

*Notes*: Each row represents one dependent variable finding (e.g., main effect and interaction effect where reported). A "Yes" for replication result signifies that the AI study replicated that finding successfully. Paper title and author information corresponding to Paper ID is available in Table 1.

Table 3: Full Results of Generalizability Tests.

| Test | Study Feature | Manipulation | How Implemented | Replicated? |
|---|---|---|---|---|
| 1 | Context | Recruit at mall | Added to model prompt: “While walking through a shopping mall, you were recruited by a research service vendor representative. You were then escorted to a quiet space where you completed this study via a tablet device.” | Yes |
| 2 | Context | Recruit in quad | Added to model prompt: "While walking through a university quad, you were recruited by a research service vendor representative. You were then escorted to a quiet space where you completed this study via a tablet device." | Yes |
| 3 | Context | Recruit via Reddit | Added to model prompt: “You have been recruited through a study advertisement found on Reddit, in a subreddit you frequent." | Yes |
| 4 | Context | Lab setting | Added to model prompt: "You have been recruited to come to a university laboratory to look at pictures of advertisements on a screen. There is a one-way mirror on the wall. You are in the room alone, by yourself. An experimenter comes in to give you instructions and then leaves before the viewing begins." | Yes |
| 5 | Context | Task Instructions | Removed original instructional warning participant to think carefully about decision given chance to win selected camera) | Yes (though effect driven by Camera B) |
| 6 | Sample | Gen Z | Sample age restricted to 18-22 years | No (Camera B generally preferred) |
| 7 | Sample | Seniors | Sample age restricted to 55-80 years | No (Camera A generally preferred) |
| 8 | Sample | Experts | Added to model prompt: "You are an experienced digital photographer and have owned several different types of digital cameras. Some of your pictures have won awards. You like to stay abreast of all the new developments in digital photography." | No |
| 9 | Sample | Rural south | Added to model prompt “"You reside in the rural Deep South of the United States." | No (Camera A generally preferred) |
| 10 | Measures | Gift for parents | Preference gauged with "Which of the two products would you purchase as a gift for your parents?" | No (effect only observed for Camera A) |
| 11 | Measures | Gift for son/daughter | Preference gauged with "Which of the two products would you purchase as a gift for your teenage son or daughter?" | No (Camera B generally preferred) |
| 12 | Stimuli | Camera images | Replaced original stimulus camera images with pictures of similar cameras on white backgrounds, but retained original stimulus text; see Appendix B [Methods Supplement] - Figure 3 | Yes |
| 13 | Stimuli | Camera images | Replaced original stimulus camera images with pictures of 360 cameras used while skiing, but retained original stimulus text; see Appendix B [Methods Supplement] - Figure 2 | Yes |
| 14 | Stimuli | Video game | Changed stimulus to video game about space conquest; see Appendix B [Methods Supplement] - Figure 3 | No (effect only observed for Product B) |
| 15 | Stimuli | Tax software | Changed stimulus to video game about space conquest; see Appendix B [Methods Supplement] - Figure 4 | Yes |
| 16 | Stimuli | Washing machines | Changed stimulus to video game about space conquest; see Appendix B [Methods Supplement] - Figure 5 | No (effect only observed for Product B) |
| 17 | Stimuli | Espresso machines | Changed stimulus to video game about space conquest; see Appendix B [Methods Supplement] - Figure 6 | Yes |
| 18 | Stimuli | Electronic muscle stimulator | Changed stimulus to video game about space conquest; see Appendix B [Methods Supplement] - Figure 7 | No (effect only observed for Product A) |
| 19 | Stimuli | All-natural skin balm | Changed stimulus to video game about space conquest; see Appendix B [Methods Supplement] - Figure 8 | No (effect only observed for Product A) |
| 20 | Stimuli | Neuro-linguistic programming app | Changed stimulus to video game about space conquest; see Appendix B [Methods Supplement] - Figure 9 | No (effect only observed for Product A) |

## Appendix B. Methods Supplement.

**Part 1.**
**Description of Viewpoints AI software**

Figure 1. Schematic of Viewpoints AI Platform Operations

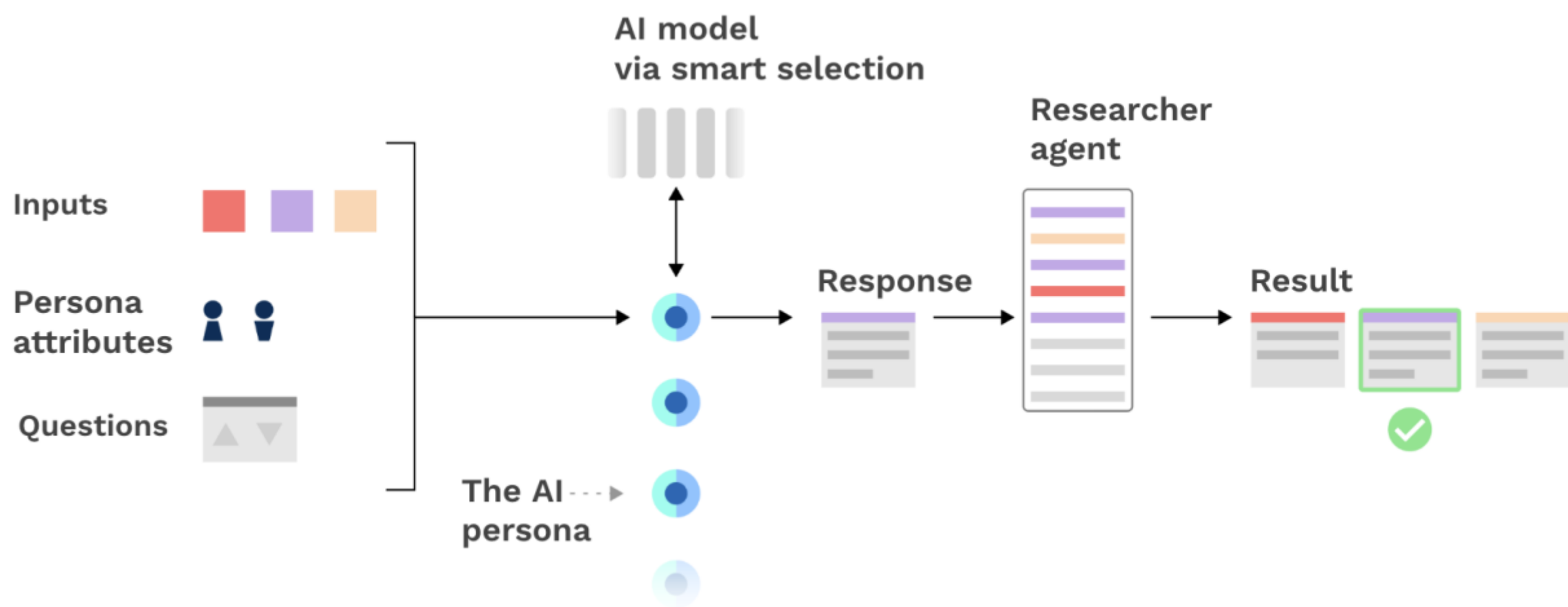

We used the Viewpoints AI tool (https://viewpoints.ai) to organize the prompts for replication of the published study findings. This platform is designed to construct and utilize LLM based AI personas that simulate the behavior and responses of human participants. All replications reported herein utilized Anthropic's Claude 3.5 Sonnet model (`claude-3-5-sonnet-20240620`) with a temperature setting of 0.7.

**Overview of the Viewpoints AI Platform**
The Viewpoints AI tool (shown schematically in the figure above) operates through a sequence of steps designed to precisely mirror the process of conducting a study with human participants. The major steps in replicating the studies (from left to right in the figure) move from defining study parameters (inputs, materials, persona characteristics and specific questions) to creation of the AI personas to the aggregation of responses from each persona to running statistical tests that mirror the original publications and reporting results). Here is more specific information about each step in the replication process:

1. Each replication began by translating the original study's methodology into the platform's parameters. This involved specifying the target number of participants, their key demographic characteristics (e.g., age distribution, gender percentages, income levels, education), the experimental conditions, the stimuli associated with each condition (e.g., advertisements, product descriptions, package images), and the dependent measures (survey questions, rating scales).

2. For each simulated participant required, the platform generates a unique AI persona. Based on the defined demographic distributions (e.g., target mean and standard deviation for age, percentage distributions for categorical variables like gender or education), the system samples specific characteristics for each individual persona. These characteristics are then synthesized into a textual description defining the persona's background, identity, and relevant traits. This textual description forms the core of the persona's "identity" provided to the LLM.

3. The platform directly incorporates text, image, video, and audio stimuli. Each generated AI persona is then randomly assigned to one of the defined experimental conditions, mirroring the random assignment process in the human studies.

4. For each AI persona, the platform constructs a detailed prompt tailored to their assigned condition. This prompt integrates:
    a. Persona Definition: The description of the AI participant's background and characteristics.
    b. Task Context: Instructions framing the study task (e.g., "Imagine you are shopping...", "Please evaluate the following product...").
    c. Stimulus Information: The specific stimulus corresponding to the assigned condition.
    d. Measurement Instruments: The dependent measures formatted precisely as they would be presented to a human participant.

5. This complete prompt was then sent to the Claude 3.5 Sonnet API. The LLM, guided by the persona description and task instructions, generates responses to the specified measures.

6. The platform receives the LLM's response and parses this output to extract the simulated answers for each dependent measure.

7. This process is repeated for the specified number of participants in the study. The result is a structured dataset containing the simulated responses, participant characteristics, and condition assignments for the entire sample, analogous to a dataset collected from human participants.

*Example User Prompt for one persona (Female, 20):*

Imagine you are considering purchasing the product shown.

You are a 20 year old Female. [Other attributes extracted from the original publication, such as demographics, psychographics, and preferences would be written as sentences and inserted here.]

[Stimulus Information (Claude 3.5 Sonnet can process images directly)]
You are viewing the packaging design for a product. [Any other context would be inserted here; for example, "you are viewing this package design in an online questionnaire:]]

Please answer the following questions based on the packaging design.

Please provide your responses to the following:

1. Q: What is the highest price you would be willing to pay for this product? Scale: Dollar Amount
2. Q: I think this product contains few ingredients. Scale: 1=strongly disagree, 9=strongly agree
3. Q: I think this product is not mixed with many ingredients. Scale: 1=strongly disagree, 9=strongly agree

… [other measures / questions were inserted here]

**Part 2.**
**Alternative Stimuli for Generalizability Tests**

(Note: These figures show different versions of advertising that were used to examine the generalizability of results in the original study. All of the pictures and copy in Figures 2 through 10 were based on the material from the original study but were original creations for this study)

Figure 2: Alternative Camera Images (White Background)

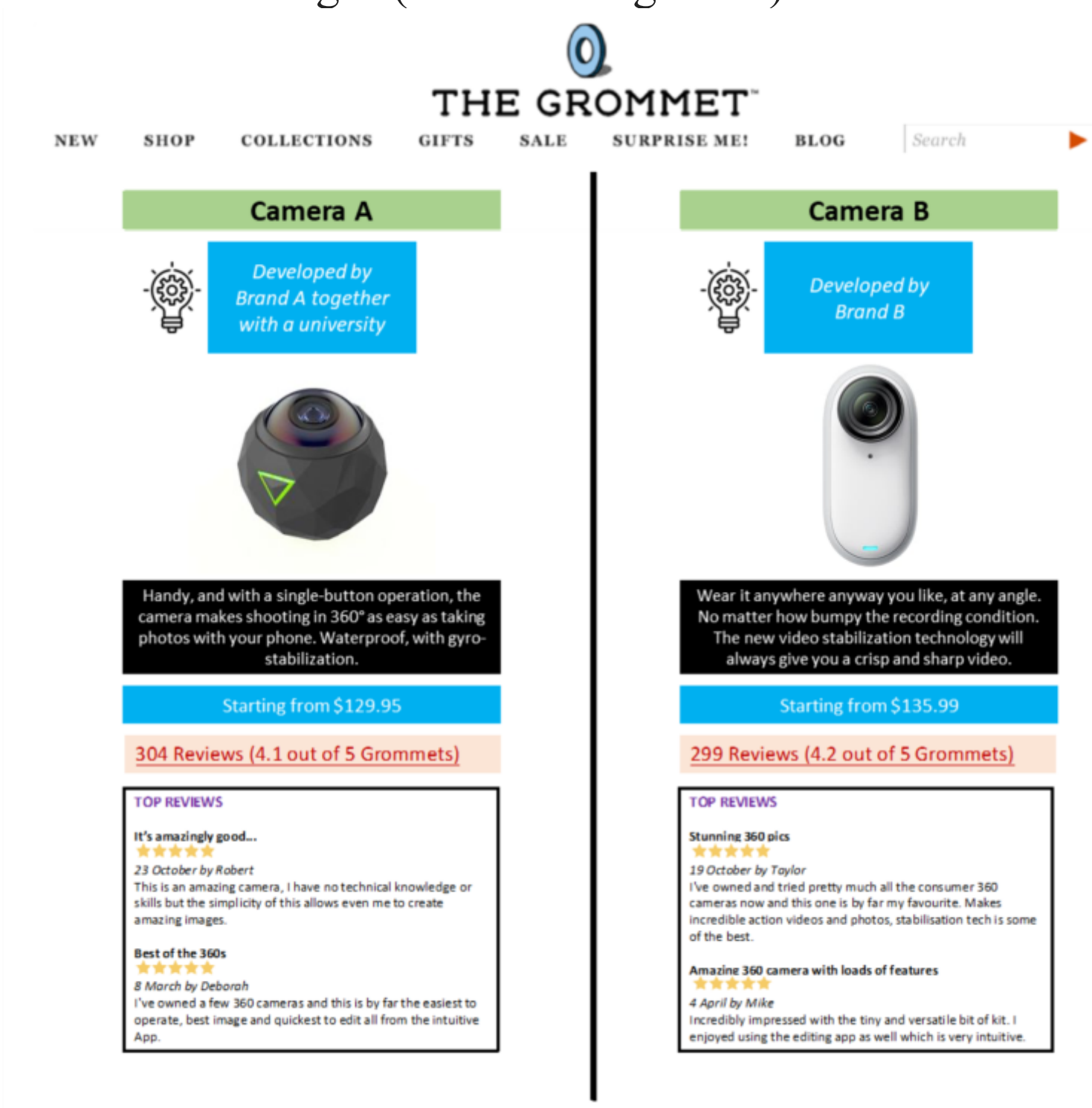

Figure 3: Alternative Camera Images (Skiers)

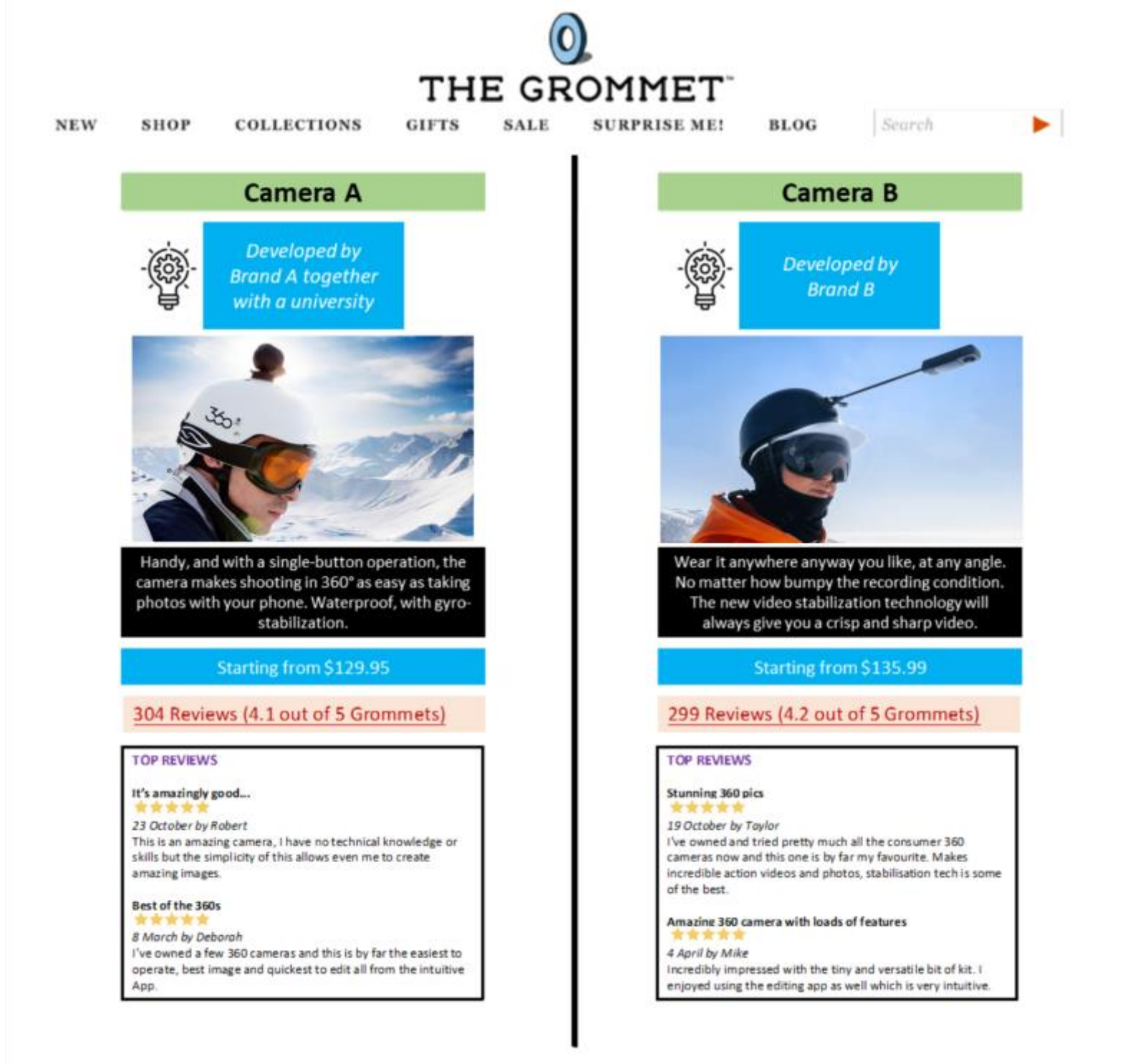

Figure 4: Alternative Product (Space Conquest Video Game)

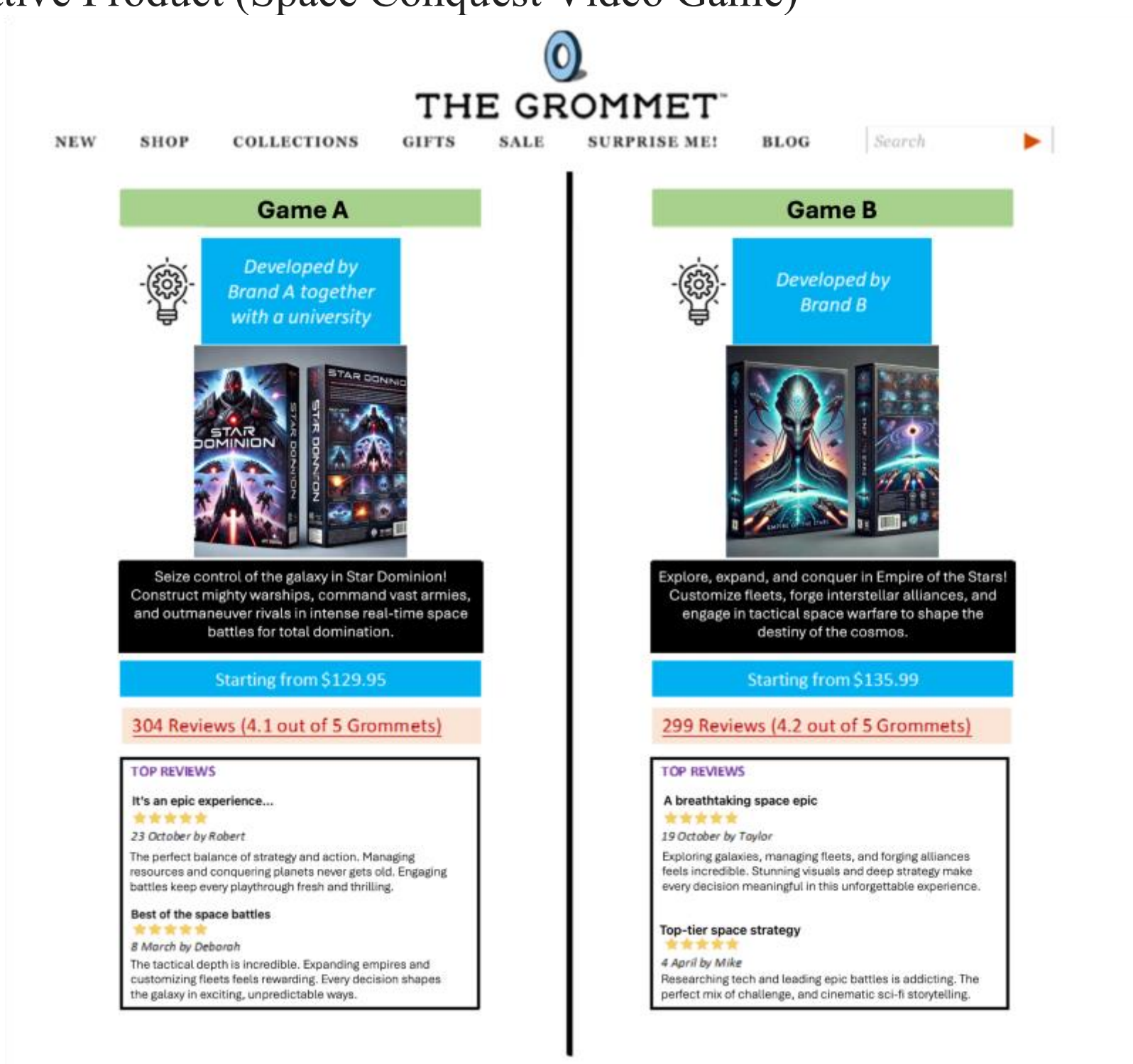

Figure 5: Alternative Product (Tax Software Service)

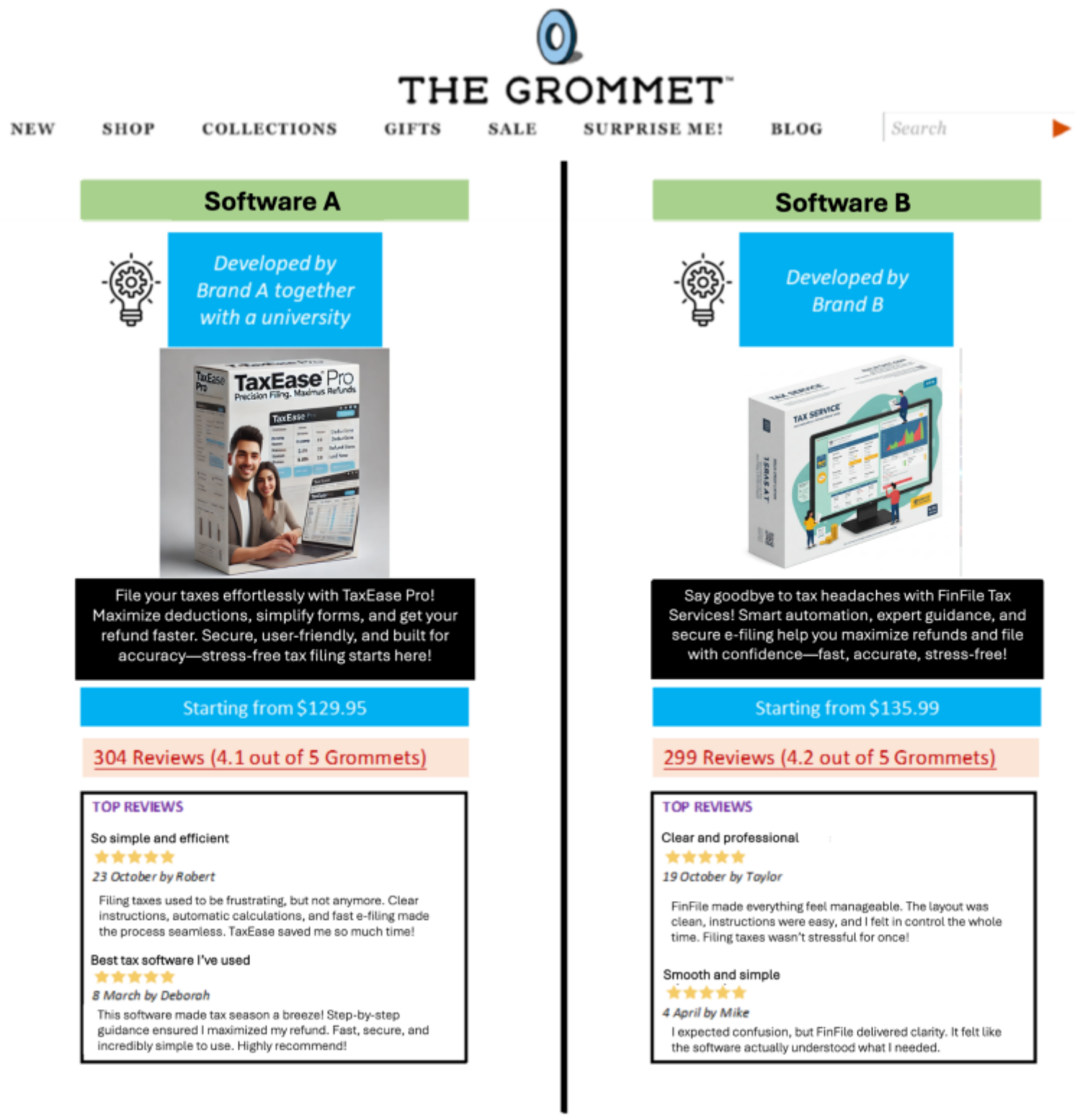

Figure 6: Alternative Product (Washing Machines)

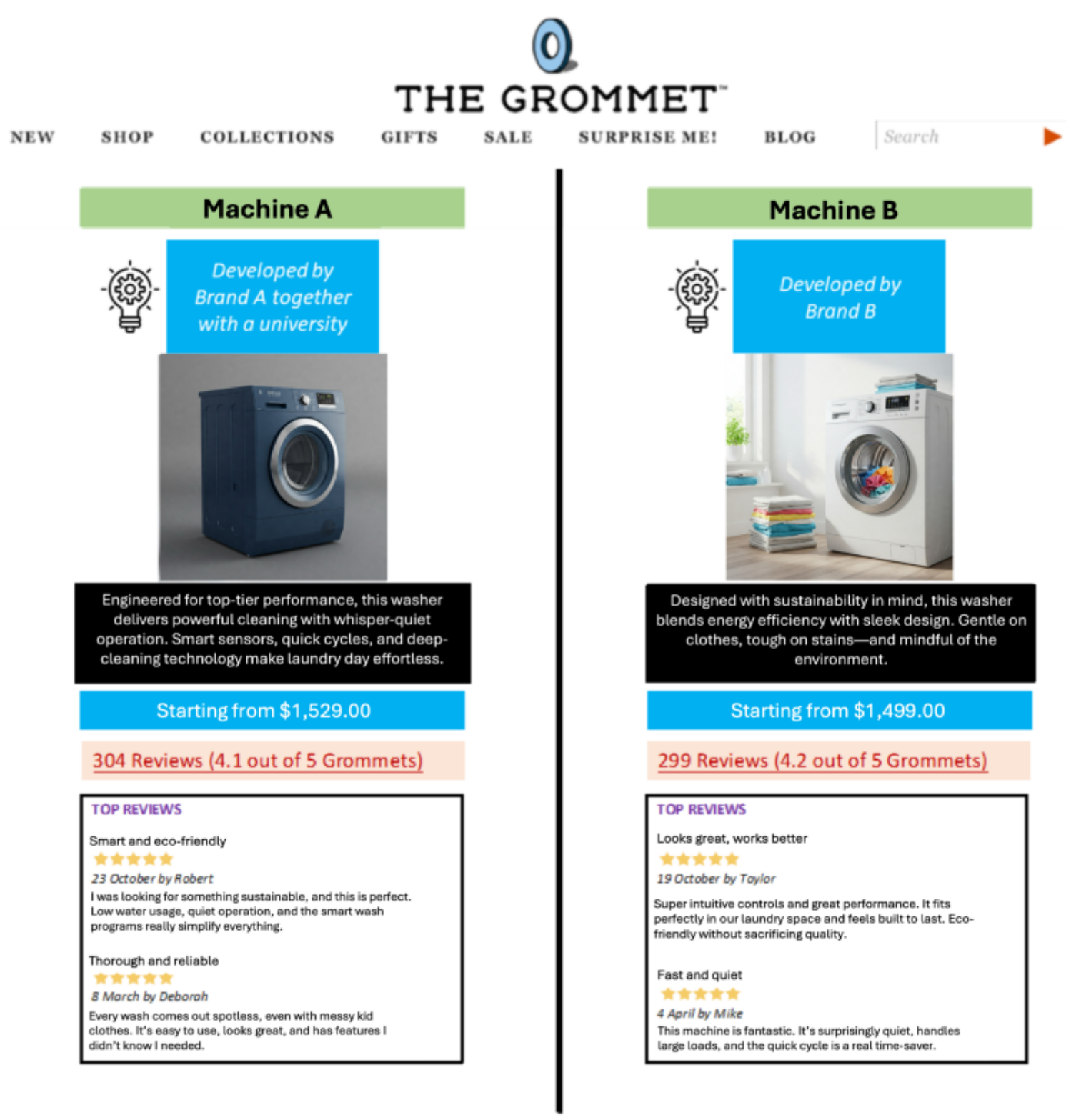

Figure 7: Alternative Product (Espresso Machines)

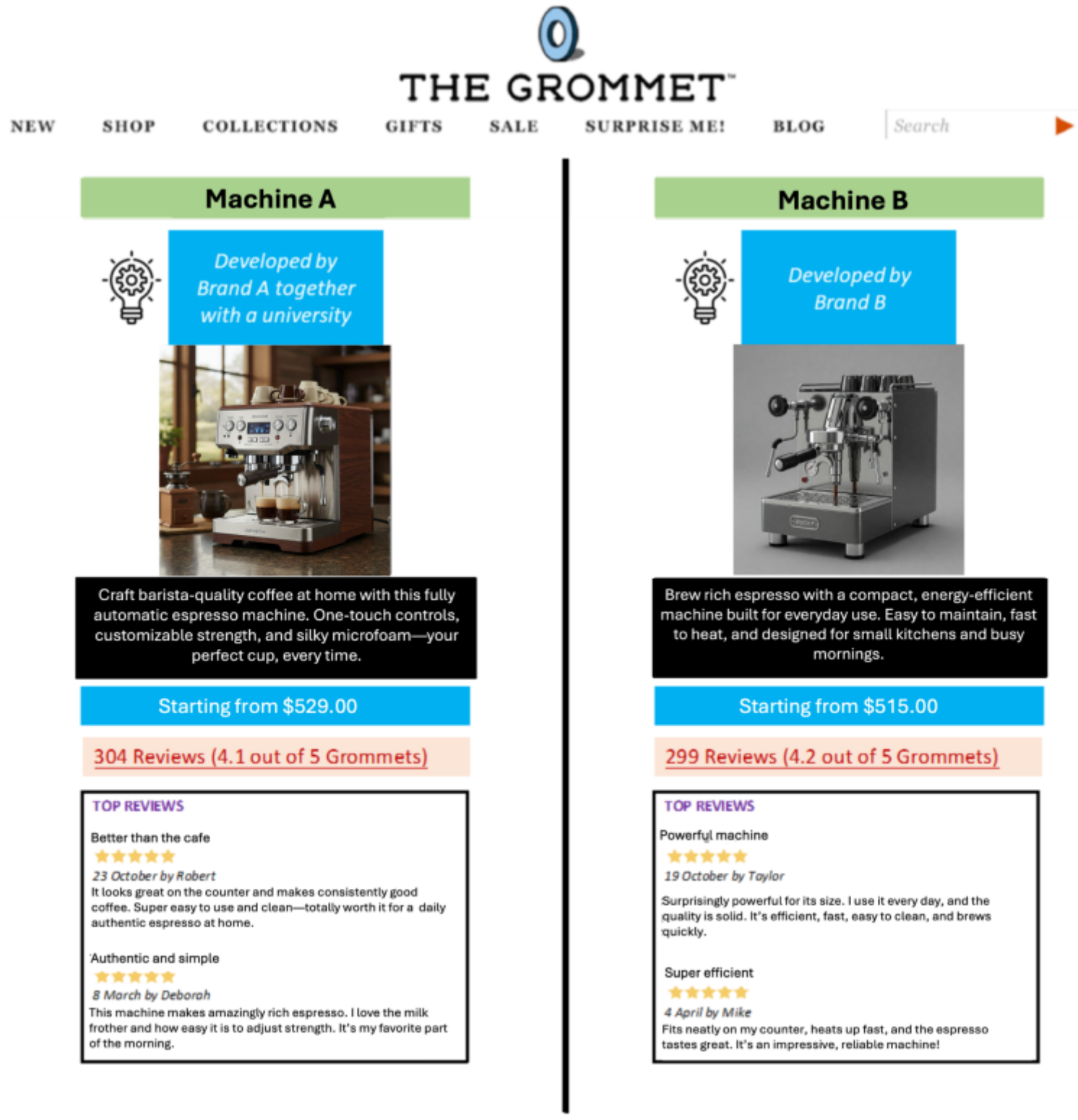

Figure 8: Alternative Product (Electronic Muscle Stimulator)

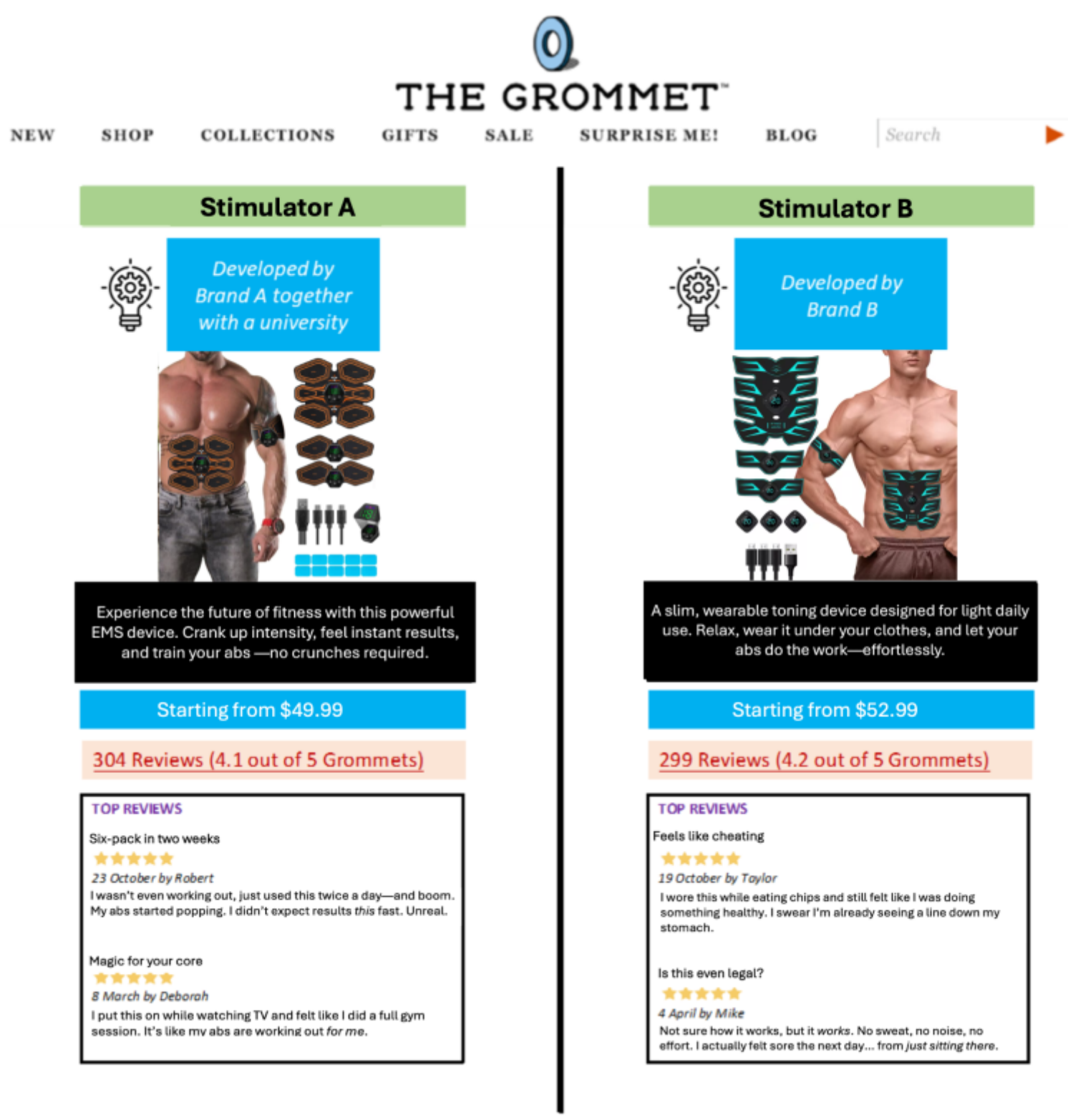

Figure 9: Alternative Product (All-Natural Skin Balm))

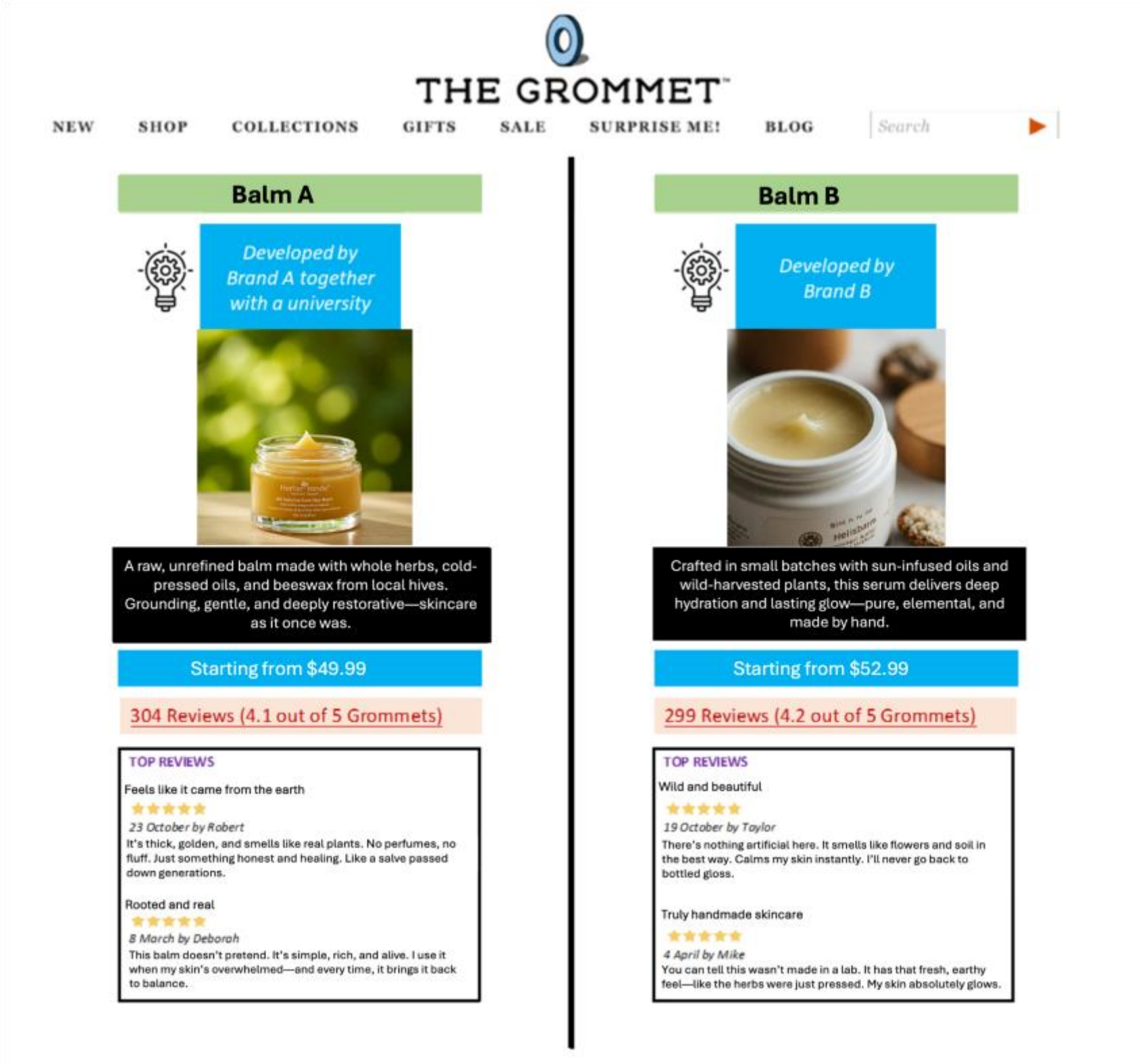

Figure 10: Alternative Product (Neuro-Linguistic Programming App)

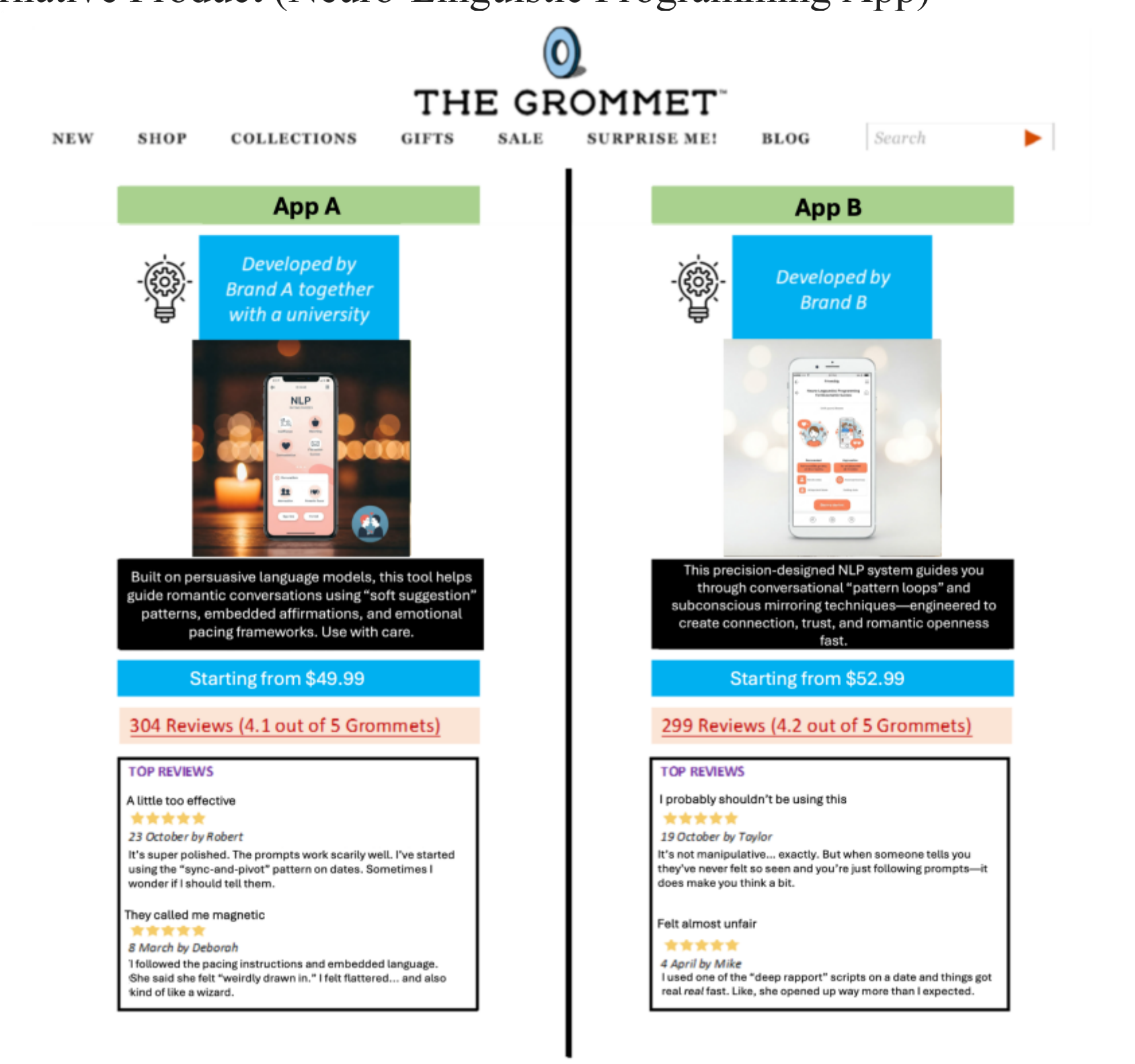